\documentclass{article}
\usepackage{arxiv}

\usepackage[normalem]{ulem}
\usepackage{multicol}

\usepackage[utf8]{inputenc}
\usepackage[T1]{fontenc}
\usepackage{hyperref}
\usepackage{url}
\usepackage{booktabs}
\usepackage{amsfonts}
\usepackage{subcaption}
\usepackage{color}
\usepackage{lipsum}
\usepackage{amsthm}
\usepackage{indentfirst}
\usepackage{graphicx}
\usepackage{subcaption}
\usepackage{epstopdf}
\usepackage{fourier}
\usepackage{bm}
\usepackage{bbm}
\usepackage{mathtools}
\usepackage{latexsym,enumerate}
\usepackage{ulem}
\usepackage{amsmath}

\usepackage{soul}
\usepackage{todonotes}

\newcommand{\BEA}{
\begin{eqnarray}}
  \newcommand{\EEA}{
\end{eqnarray}}
\newcommand{\bef}{{\bf f}}

\newcommand{\bu}{{\bf u}}

\newcommand{\bx}{{\bf x}}
\newcommand{\by}{{\bf y}}

\newcommand{\bU}{{\bf U}}
\newcommand{\bV}{{\bf V}}
\newcommand{\bX}{{\bf X}}

\newcommand{\norm}[1]{\left\lVert#1\right\rVert}

\newcommand{\vtf}{\boldsymbol{\varphi}}
\newcommand{\vtS}{\boldsymbol{\Sigma}}
\newcommand{\BR}{\mathbb{R}}

\newcommand{\lreg}{{\lambda_{\mathrm{reg}}}}

\newcommand{\comment}[1]{}

\title{Learning solution operator of dynamical systems with diffusion maps kernel ridge regression}

\author{
  Jiwoo Song, Daning Huang\footnote{Corresponding author: daning@psu.edu}\\
  Department of Aerospace Engineering \\
  The Pennsylvania State University, University Park, PA 16802, USA\\
  \texttt{jzs6565@psu.edu, daning@psu.edu} \\
  \And
  John Harlim \\
  Department of Mathematics, Institute for Computational and Data Sciences \\
  The Pennsylvania State University, University Park, PA 16802, USA\\
  \texttt{jharlim@psu.edu} 
}

\begin{document}

\maketitle

\begin{abstract}
In this work, we propose a simple kernel ridge regression (KRR) framework with a dynamic-aware validation strategy for long-term prediction of complex dynamical systems. By employing a data-driven kernel derived from diffusion maps, the proposed Diffusion Maps Kernel Ridge Regression (DM-KRR) method implicitly adapts to the intrinsic geometry of the system's invariant set, without requiring explicit manifold reconstruction or attractor modeling, procedures that often limit predictive performance. Across a broad range of systems, including smooth manifolds, chaotic attractors, and high-dimensional spatiotemporal flows, DM-KRR consistently outperforms state-of-the-art random feature, neural-network and operator-learning methods in both accuracy and data efficiency. These findings underscore that long-term predictive skill depends not only on model expressiveness, but critically on respecting the geometric constraints encoded in the data through dynamically consistent model selection. Together, simplicity, geometry awareness, and strong empirical performance point to a promising path for reliable and efficient learning of complex dynamical systems.
\end{abstract}

\section{Introduction}

In many scientific and engineering domains, ranging from multibody mechanics and fluid dynamics to circuit simulation and power systems, the governing dynamics are often modeled as high-dimensional systems of nonlinear ordinary differential equations (ODEs). Although these ODE models are often derived from first principles, they inevitably suffer from modeling errors due to incomplete or imperfect understanding of the underlying physics. A broad range of approaches has been proposed to mitigate these discrepancies, with successful applications reported across various complex systems, including recent advances that leverage machine learning.

Broadly speaking, there are two common approaches in literature.
The fully data-driven approach ignores the first-principle model and learns the entire solution operator or the vector field of the ODEs directly from observed data (see, e.g.,~\cite{wang2005gaussian,williams2015data,brunton2016discovering,vlachas2018data,pathak2018model,hamzi2021learning,huang2025learning}).
In contrast, the partially data-driven approach retains the first-principles model and learns only the model-error components that correct the underlying dynamics (see, e.g.,\cite{kaheman2019learning,lin2021data,harlim2021machine,levine2022framework,venegas2025a}). In this study, we focus on the fully data-driven setting, although the proposed framework could, in principle, also be adapted for learning model-error components in partially data-driven formulations.

Despite the typically high-dimensional  phase space representation, the dynamics typically remain confined to a much lower-dimensional region, especially over long time horizons. For regular systems, this region forms a smooth manifold embedded in the phase space. For chaotic systems, the long-term dynamics typically forms a rough structure known as fractal (or strange) attractor. These long-term structures are known as the \emph{forward invariant sets}.

Accurately predicting system evolution while respecting the constraints imposed by these invariant sets is critical in many real-world applications, e.g., weather forecasting, nuclear fusion control, and embodied intelligence. Since invariant sets are rarely known explicitly, a practical challenge is to identify an effective dictionary of candidate functions (or features) for learning the underlying dynamics. Naive choices of linear or nonlinear combinations of functions in the ambient phase-space coordinates (e.g., SINDy \cite{brunton2016discovering}) lead to a combinatorial explosion in the number of candidate features, rendering the problem computationally prohibitive and statistically unstable.

Many methods have been proposed to identify low-dimensional representations of dynamical systems, ranging from linear techniques such as PCA (Principal Component Analysis) \cite{jolliffe2011principal} and DMD (Dynamic Mode Decomposition) \cite{williams2015data} and its variants, to nonlinear approaches based on autoencoders \cite{Linot2020,Maulik2021,vlachas2018data,Regazzoni2024}. A limitation of PCA and DMD is that they sometimes fail to produce a finite, low-dimensional representation capable of capturing the full dynamics, even when the underlying invariant set itself is finite-dimensional. Moreover, algorithms in the DMD family approximate the Koopman operator and therefore require additional treatment to avoid spectral pollution. A promising algorithm to avoid spectral pollution in the Koopman spectral approximation is ResDMD \cite{colbrook2024rigorous}, which we include as a baseline in one of our test cases. As for autoencoders, it is found that the latent space only provides a global parametrization of the invariant set, whose dimension is typically larger than the intrinsic dimension of the invariant set, and the dynamics are not guaranteed to learn the exponential maps (for discrete-time model) or vector fields (for continuous-time model) while satisfying the invariant set constraint \cite{huang2025learning}. As a result, the predicted trajectories may deviate from the invariant set, limiting the long-term prediction accuracy, as we shall see in several numerical examples with neural network (NN) baselines.
Meanwhile, there is indeed an autoencoder-based approach, CANDyMan, that can uncover the invariant set when it is a smooth manifold \cite{Floryan2022}. In such a case, the estimated dynamics are guaranteed to evolve on the manifold. However, this approach cannot be extended to the case where the invariant set is fractal.
In addition, common to NN-based approaches, the computational effort to train the models is not negligible and the hyperparameter tuning process can be difficult.

A promising approach with solid theoretical foundations for mitigating the curse of dimensionality is to exploit the “kernel trick,” which circumvents the explicit evaluation of high-dimensional feature spaces. For dynamical systems evolving on smooth manifolds (i.e., those whose forward invariant sets form smooth submanifolds of the phase space), the Geometric Constrained Multivariate Kernel Ridge Regression (GMKRR) method was recently introduced to approximate the governing vector field of the ODEs \cite{huang2025learning}. GMKRR is paired with a Normal-Correction integration scheme that ensures time integration remains consistent with the manifold constraint. Benchmark studies have shown that GMKRR outperforms several state-of-the-art NN-based methods \cite{Floryan2022,Regazzoni2024}, achieving superior long-term predictive accuracy while reducing training costs by 3–4 orders of magnitude. Despite these promising results, GMKRR remains limited by the accuracy of tangent-bundle approximation and cannot be directly applied to systems whose invariant sets exhibit fractal structure.

In this paper, we address the challenge of long-term prediction for dynamical systems evolving on unknown invariant sets using a scalar-valued, data-driven kernel constructed from the diffusion maps (DM) algorithm \cite{coifman2006diffusion,berry2016variable}. Specifically, if we denote the solution operator of a dynamical system as $\vtf_t =(\varphi_{1,t},\ldots,\varphi_{n,t}): \mathbb{R}^n \to \mathcal{M} \subset \mathbb{R}^n$, which maps any initial condition $\bx(0)$ to $\bx(t)$, where $\mathcal{M}$ denotes the forward-invariant set, then we consider either a \emph{direct} estimator, 
\begin{equation}
x_j(t_{i+1}) =\varphi_{j,\Delta t}(\bx(t_i)) \approx f_{j}(\bx(t_i)) = k_{\epsilon,N}(\bx_i, X)\boldsymbol{\alpha}_j     
\end{equation}
or a \emph{skip-connection} estimator, 
\begin{equation}
x_j(t_{i+1}) - x_j(t_{i}) =  \varphi_{j,\Delta t}(\bx(t_i)) - x_j(t_i) \approx f_j(\bx(t_i)) = k_{\epsilon,N}(\bx_i, X)\boldsymbol{\alpha}_j,    
\end{equation}
for each component $j=1,\ldots,n$. Here,  $k_{\epsilon,N}:\mathbb{R}^n \times \mathbb{R}^n \to \mathbb{R}$ denotes the data-dependent DM kernel with a lengthscale parameter $\epsilon$, $X=\{\bx_i =\bx(t_i)= (x_1(t_i),\ldots, x_n(t_i))\}_{i=1,\ldots, N}$ denotes the training dataset with $t_i =i \Delta t$,  and $\boldsymbol{\alpha}_j$ denotes the $N$-dimensional regression coefficient for the $j$th component. While \emph{direct} estimator can be used in general, we notice substantial improvement with \emph{skip-connection} estimator when the underlying ODE systems are not numerically stiff.
In this study, we consider a Kernel Ridge Regression solution with DM kernel, coined as Diffusion Maps Kernel Ridge Regression (DM-KRR), where the regression coefficient, $\boldsymbol{\alpha}_j = \boldsymbol{\alpha}_j(\epsilon,\lreg)$, depends on the shape parameter, $\epsilon$, and ridge regression parameter, $\lreg$. We will show that careful choice of validation metric to determine the model hyperparameters $(\epsilon,\lreg)$ is essential for achieving optimal predictive performance. While the conclusion here resonates with the conclusion from the Kernel Flow (KF) approach \cite{hamzi2021learning},
the present approach differs significantly from KF. Particularly, KF parameterizes the flow map using a family of kernels and selects among them via validation over the same loss-square error, we employ a fixed, data-driven kernel derived the from diffusion maps and validate by comparing the prediction error of iterating the approximate model over a time window. 

Beyond its simplicity and strong empirical performance, our use of kernel methods is also theoretically motivated. If the underlying dynamical system is ergodic with invariant measure 
$\mu$ supported on a compact set $A \subset \mathcal{M}\subset \mathbb{R}^n$, then the Reproducing Kernel Hilbert Space (RKHS) induced by any continuous kernel $k$ on a compact set $A$ is universal for approximating $L^2(\mu)$ functions, even when $A$ lacks smoothness. Thus, in principle, one may employ any such kernel, e.g., Gaussian or Mat\'{e}rn. Additionally, when the forward invariant set $\mathcal{M} \supset A$ is a smooth $d$-dimensional sub-Riemannian manifold and the invariant measure is uniformly distributed with a compact support that lies in the invariant set, the diffusion maps (DM) kernel is a natural choice as it approximates the heat kernel in $L^\infty$-sense over the dataset as shown in \cite{dunson2021spectral}. 
In effect, the DM kernel encodes the intrinsic geometry of the invariant set, allowing the learned dynamics to adapt seamlessly to its true intrinsic dimensionality.

The remainder of this paper is organized as follows. In Section~\ref{method}, we give a quick overview of the kernel ridge regression for our setup and introduce a dynamic-aware validation strategy to select the model hyperparameters.
In Section~\ref{DMkernel}, 
we introduce the diffusion maps kernel and discuss some theoretical results that motivate this choice of kernel. In Section~\ref{results}, we demonstrate the numerical performance of DM-KRR over a range of dynamical systems whose forward invariant sets may be smooth or non-smooth and are embedded in ambient spaces ranging from dimension 
$3$ to $10^5$. In Section~\ref{summary}, we close the paper with a short summary and discussion. We supplement the paper with four appendices. Appendix~\ref{sec:si_heur} gives a heuristic estimate to assist the validation procedure. 
The remaining three appendices (\ref{sec:si_case1}-\ref{sec:si_case5}) document additional details that correspond to the numerical examples.

\comment{In the next section, we demonstrate the advantages of the DM kernel over Gaussian kernels across a range of dynamical systems whose forward invariant sets may be smooth or non-smooth and are embedded in ambient spaces ranging from dimension 
$3$ to $10^5$. In several examples, we also compare against autoencoder-based approaches and ResDMD.}

\section{Kernel Ridge Regression with Dynamic-Aware Validation Strategy}\label{method}

In this section, we give a short overview of kernel ridge regression (KRR) and a dynamic-aware validation strategy to select the model hyperparameters.

\subsection{Kernel Ridge Regression}\label{sec:krr}

Given a set of training data $S  = (X,Y)$, where $X = [\bx_1,\ldots,\bx_N]$ and $Y =[\by_1,\ldots,\by_N]$, the KRR is to find a map $\mathbf{f}: \mathbb{R}^n \mapsto \mathbb{R}^n$ of the following form, 
\begin{equation}
f_j = k_\epsilon(\cdot, X) \boldsymbol{\alpha}_j = \sum_{i=1}^N  k_\epsilon (\cdot, \bx_i)\alpha_{j,i},\label{fs}
\end{equation}
where $k_\epsilon:\mathbb{R}^n \times \mathbb{R}^n \to \mathbb{R}$ is a scalar-valued kernel. In the above notation, $k_\epsilon(\cdot,X):\mathbb{R}^n \to \mathbb{R}^N$, where we evaluate the kernel's second input on all data points in $X$.  

The classical kernel ridge regression (KRR) solution to this modeling problem is to solve 
\begin{equation}
\min_{ \boldsymbol{\alpha}_j} L_j(\boldsymbol{\alpha}_j)  := \min_{ \boldsymbol{\alpha}_j}\sum_{i=1}^N \left((Y_{ji} - f_j(\bx_i)\right)^2 +
\lreg \|\boldsymbol{\alpha}_j\|^2,\label{KRR}
\end{equation}
for each component $j = 1\ldots, N$, where $\lreg$ is a user-specified ridge parameter for regularization. In the notation above, we denote $Y_{ji}$ as the $(j,i)$th component of $Y$. This problem admits a unique solution with $\boldsymbol{\alpha}_j = (\mathbf{K}_N+\lreg\mathbf{I}_N)^{-1} Y_j^\top$, where $\mathbf{K}_N$ is an $N\times N$ Gram matrix with $(i,j)$-th component $k_\epsilon(\bx_i,\bx_j)$, $\mathbf{I}_N$ is an identity matrix of size $N\times N$, and $Y_j$ denotes the $j$th row of data matrix $Y$. 

From an ensemble of $J$ timeseries of length $T$, the input data for KRR is constructed as
$$
X = \left[\bx_1^{(1)},\bx_2^{(1)},\cdots,\bx_{T-1}^{(1)},\cdots,\bx_1^{(J)},\bx_2^{(J)},\cdots,\bx_{T-1}^{(J)}\right] \in \mathbb{R}^{n\times N},
$$
so $N=J\times(T-1)$. In the case of \emph{direct} solution operator, the output data is constructed as
$$
Y = \left[\bx_2^{(1)},\bx_3^{(1)},\cdots,\bx_{T}^{(1)},\cdots,\bx_2^{(J)},\bx_3^{(J)},\cdots,\bx_{T}^{(J)} \right]\in \mathbb{R}^{n\times N},
$$
In the case of \emph{skip-connection}, the procedure is the same, except that the output data is replaced as
$$
Y = \left[\Delta\bx_1^{(1)},\Delta\bx_2^{(1)},\cdots,\Delta\bx_{T-1}^{(1)}, \cdots, \Delta\bx_1^{(J)},\Delta\bx_2^{(J)},\cdots,\Delta\bx_{T-1}^{(J)}\right] \in \mathbb{R}^{n\times N},
$$
where $\Delta\bx_i^{(j)} = \bx_{i+1}^{(j)} - \bx_{i}^{(j)}.$

\subsection{Dynamic-Aware Validation Strategy}\label{sec:hyperparam}

The KRR formulation considered in this study has two hyperparameters: the regularization factor, $\lreg$, in the KRR, and the lengthscale hyperparameter, $\epsilon$, in both the RBF and DM kernels.  The choice of the two hyperparameters critically impacts the accuracy of the learned solution operator. 
The standard validation strategy is to find the optimal $\epsilon,\lreg$ over a range of hyperparameters $H=[\epsilon_l, \epsilon_h]\times [\lreg_{,l}, \lreg_{,h}]$, in the sense that they minimize the same loss function $L_j$ in \eqref{KRR} over a validation data set. Such a validation metric, while it is useful other application, is not appropriate for our purpose since we ultimately want to iterate $\mathbf{f}$ to predict the dynamical systems. 


For our application, we require a validation dataset consisting of a number (usually about 3-12 in our numerical experiments) of trajectories of length $N_v$. Instead of minimizing the same loss function $L_j$, we consider the following validation metrics.

For a validation trajectory of length $N_V$, $\{\bx_i\}_{i=1}^{N_V}$, suppose the rollout prediction is $\{\hat{\bx}_i\}_{i=1}^T$, where $\hat{\bx}_{i+1}=\bef(\hat{\bx}_i)$ for $i=1,2,\cdots,T-1$, with $\hat{\bx}_1=\bx_1$ and $\bf{f}$ denoting the learned solution operator for a particular choice of $\epsilon,\lreg$.  The validation metrics are defined as follows:

\begin{enumerate}  
\item \textit{Root Mean Squared Error (RMSE)}: 
For dynamical systems with smooth forward invariant sets, we consider RMSE evaluated over the prediction horizon as
\begin{equation}\label{eqn:rmse}
    \text{RMSE}= \sqrt{\frac{1}{nN_V}\sum_{i=1}^{N_V} \left\|\hat{\bx}_i-\bx_{i}\right\|_2^2}.
\end{equation}

\item \textit{Valid Prediction Time (VPT)}:
For chaotic dynamical systems, we consider the VPT, defined as the maximum duration for which the normalized trajectory error remains below a certain error tolerance $\gamma$,
\begin{equation}
    \mathrm{VPT} = \frac{1}{T_{\Lambda}} \max \left\{i|E_i\leq \gamma \text{ for all } i\leq N_V\right\},\label{eq:VPT}
\end{equation}
where $T_\Lambda = 1/\Lambda$ is the unit Lyapunov time, with $\Lambda$ as the maximal Lyapunov exponent of the system.  The normalized error at time step $i$ is
$$
E_i = \sqrt{\frac{1}{n}\sum_{i=1}^n \left(\frac{\hat{x}_{k,i}-x_{k,i}}{\sigma_k}\right)^2 },
$$
where $x_{k,i}$ is the $k$-th component of the state $\bx_i$ and $\sigma_k$ is the standard deviation of the $k$-th component of the true state over the trajectory.
\end{enumerate}

Let $E$ be either RMSE or VPT as a validation metric. We conduct the following random search strategy for the validation procedure.  It starts with a given range of hyperparameters $H=[\epsilon_l, \epsilon_h]\times [\lreg_{,l}, \lreg_{,h}]$, and the maximum number of trials.
For each trial,
\begin{enumerate}
    \item Randomly select $(\epsilon,\lreg)\in H$.
    \item Use the training dataset of size $N$ to learn the solution operator, using $(\epsilon,\lreg)$ for KRR.
    \item Using the learned KRR model, predict on the validation trajectories, and compute the average of the validation metric $E$, averaged over the number of validation trajectories.
\end{enumerate}
The model associated with the smallest validation metric is used as the final model for testing. 

The range of hyperparameters for random search may vary significantly among different problems.  While it is possible to use a generic adaptive grid search approach, to reduce the computational cost, we used a heuristic choice of the range.  First, a reference lengthscale value $\epsilon^*$ is computed using an estimation algorithm, provided in the Appendix \ref{sec:si_heur}, based on the input data for training.  Then, using $\epsilon^*$ the symmetric RBF kernel matrix is constructed and suppose its minimum eigenvalue is $\lambda_{\min}$; choose the reference regularization factor $\lreg^*=|\lambda_{\min}|$.  Subsequently, the range is chosen to be $H=[\epsilon^*\Delta \epsilon, \epsilon^*/\Delta \epsilon]\times [\lreg^*, \lreg^*/\Delta \lreg]$, where, unless otherwise stated, $\Delta \epsilon=10^{-2}, \Delta \lreg=10^{-4}$.

\section{Diffusion Maps Kernel} \label{DMkernel}

Although the KRR model above can be used with any kernel, the quality of the model prediction depends crucially on the choice of kernel being used. It is well known that choosing a kernel is typically problem dependent since  each kernel is associated to a class of hypothesis models that may or may not be appropriate for some applications.

\comment{
In principle, any kernel can be used as a KRR model, e.g., the following Gaussian radial basis function (RBF) kernel, \begin{equation}\label{eqn_rbf}
\tilde{k}_{\epsilon}(\bx,\by) = \exp\left(-\frac{\|\bx-\by\|^2}{4\epsilon}\right),
\end{equation}
can be used to solve the KRR problem above.}

In Subsection~\ref{subsection-DM}, we introduce a data-driven scalar-valued kernel induced by the Diffusion Maps (DM) algorithm \cite{coifman2006diffusion}. Besides the strong empirical performances across various examples to be reported in Section~\ref{results}, we also provide a theoretical discussion in subsection~\ref{theory} that hinges on a conjecture that may explain why the DM kernel is an appropriate choice for our applications.

\subsection{Formulation of the DM Kernel}\label{subsection-DM}

The Diffusion Maps algorithm constructs a transition matrix based on a Markov chain representation of heat diffusion on the data points. 
Closer points are more likely to transition to each other, and the transition matrix is constructed to reflect this. 
This algorithm has multiple possible normalizations to remove biases induced by the sampling distribution.
Given a dataset $X = \{\bx_1,\ldots, \bx_N\}$ (drawn i.i.d.~with sampling measure $\mu$) lies on a $d$-dimensional, smooth, compact, Riemannian submanifold of $\mathcal{M} \subset\mathbb{R}^n$, our chosen normalization corresponds to a Graph Laplacian matrix that converges pointwisely and spectrally to the semigroup solution of the Laplace-Beltrami operator \cite{coifman2005geometric, coifman2006diffusion, berry2016variable,dunson2021spectral,calder2022lipschitz,Peoples2025}. In this case, the eigenfunctions of the Laplace-Beltrami operator form a complete basis for the integrable functions on the manifold, and hence are arguably effective in representing the solution operators in this study. 

Our main idea is to consider a symmetric matrix that is diagonally similar to the Graph Laplacian matrix constructed by DM as a Gram matrix for Kernel Ridge Regression.
While the theoretical justification is restricted to smooth manifold hypothesis, we employ the kernel on arbitrary datasets $X \subset \mathcal{M}$, where $\mathcal{M}\subset \mathbb{R}^n$ denotes a compact forward invariant set that is not necessarily a smooth Riemannian manifold.

For simplicity of discussion, we only consider the fixed bandwidth diffusion maps algorithm with the appropriate normalizations to generate our desired Laplace-Beltrami operator; although a symmetric kernel induced by the variable bandwidth diffusion maps \cite{berry2016variable} or the symmetrized kernel in \cite{Peoples2025} could also be used in principle.
To construct our kernel, we perform two normalization stages,
on the symmetrically weighted Gaussian RBF kernel,
\begin{equation}\label{eqn_rbf}
\tilde{k}_{\epsilon}(\bx,\by) = \exp\left(-\frac{\|\bx-\by\|^2}{4\epsilon}\right).
\end{equation}
First, the ``right'' normalization in DM is to remove the bias induced by non-uniform distribution of the data and is defined as,
\[
  \hat{k}_{\epsilon}(\bx,\by) =  q_{\epsilon}^{-1}(\bx)\tilde
  k_{\epsilon}(\bx, \by)q_{\epsilon}^{-1}(\by),
\]
where $q_\epsilon(\bx) = \int_{\mathcal{M}} \tilde k_{\epsilon}(\bx, \by)q(\by)\,d\text{Vol}(\by)$, with $q$ denoting the sampling density of the dataset ($d\mu = qd\text{Vol}$). We note that $\hat k_{\epsilon}$ is symmetric and positive definite. The second normalization is with respect to $\hat{q}_{\epsilon}(\bx) = \int_{\mathcal{M}} \hat{k}_{\epsilon}(\bx,\by) q(\by)d\text{Vol}(\by)$ to attain the Markov transition kernel,
\[
  k^{DM}_{\epsilon}(\bx,\by) =\hat{q}_{\epsilon}^{-1}(\bx)  \hat{k}_{\epsilon}(\bx,\by),
\]
of a reversible Markov chain on $\mathcal{M}$.

Numerically, since the sampling density $q$ is unknown, we approximate $k_{\epsilon}$ with $k_{\epsilon,N}$ obtained through the following algebraic manipulation,
\begin{equation}
q_{\epsilon,N}(\bx) = \frac{1}{N}\sum_{j = 1}^N \tilde k_\epsilon(\bx, \bx_j),\quad
\hat k_{\epsilon,N}(\bx, \by) = \frac{\tilde k_{\epsilon}(\bx,
\by)}{q_{\epsilon,N}(\bx) q_{\epsilon,N}(\by)},\quad
\hat{q}_{\epsilon,N}(\bx) = \frac{1}{N}\sum_{j = 1}^N \hat
k_{\epsilon,N}(\bx, \bx_j), \label{MCstep}
\end{equation}
and finally arrive at the approximate transition kernel, 
\begin{equation}
    k_{\epsilon,N}^{DM}(\bx,\by) = \hat{q}_{\epsilon,N}^{-1}(\bx) \hat{k}_{\epsilon,N}(\bx,\by).\label{transkernel} 
\end{equation}
We define the {\bf DM kernel} to be the following symmetric kernel,
\begin{equation}
    k_{\epsilon,N}(\bx,\by) = \hat{q}_{\epsilon,N}^{1/2}(\bx) k^{DM}_{\epsilon,N}(\bx,\by)q_{\epsilon,N}^{-1/2}(\by) = \hat{q}_{\epsilon,N}^{-1/2}(\bx) \hat{k}_{\epsilon,N}(\bx,\by) \hat{q}_{\epsilon,N}^{-1/2}(\by).\label{scalarkernel}
\end{equation}
Notice that this is a data-dependent kernel through the empirical average in \eqref{MCstep}.


\subsection{Reproducing Kernel Hilbert Space Corresponding to the DM Kernel}\label{theory}

In the following discussion, we identify the Reproducing Kernel Hilbert Space (RKHS) corresponding to the DM kernel in \eqref{scalarkernel}. For this discussion, we assume that $\mathcal{M}$ is compact. Define $S_k: L^2(\mu) \to \mathcal{H}$, as, 
\begin{eqnarray}
S_k f = \int_{\mathcal{M}} k(\cdot,\by) f(\by) d\mu(\by),\label{sk}    
\end{eqnarray}
where $\mathcal{H}$ is the RKHS corresponding to the reproducing kernel $k$. One can show that its adjoint $S_k^*=id: \mathcal{H} \to L^2(\mu)$ is an identity operator. Define $T_k = S_k^*S_k$ which is the same integral operator as above, except that it is mapping from $L^2(\mu)$ to $L^2(\mu)$.
For any bounded kernel, $\|k\|_{L^2(\mu)} <\infty$, the spectral theory (see Theorem 4.9.19 in \cite{debnath2005introduction}) states that $T_k$ has real eigenvalues $\lambda_0\geq  \lambda_1 \geq \ldots \searrow 0$. For Markov kernel, $\lambda_0=1$. The corresponding eigenfunctions form a Riesz basis for $L^2(\mu)$.
 The Mercer's theorem suggests that,
\begin{eqnarray}
k(\bx,\by) = \sum_{i\geq 0} \lambda_i \psi_i(\bx)\psi_i(\by),\label{Mercer}
\end{eqnarray}
where $\psi_i = \lambda_i^{-1} S_k\varphi_i$ is the continuous representation of the eigenfunction $\varphi_i$ for any nonzero $\lambda_i$. This implies that the RKHS 
\begin{equation}
\mathcal{H} = \left\{f = \sum_{i\geq 0} a_i\sqrt{\lambda_i} \psi_i, \{a_i \} \in \ell_2\right\},\label{RKHS}   
\end{equation}
is dense in $L^2(\mu)$ when $\lambda_i\neq 0$ \cite{christmann2008support}. 

In our case, the family of DM kernel  
$k_{\epsilon,N}$ in \eqref{scalarkernel} is considered with $\mu$ in \eqref{sk} being replaced by an empirical measure, $\mu_N = \frac{1}{N}\sum_{i=1}^N \delta( \bx-\bx_i)$. We denote the corresponding family of RKHS as $\mathcal{H}_{\epsilon,N}$ as in \eqref{RKHS} with eigenpairs to be determined in the following discussion. 
Let $\lambda_{\epsilon,N,i}$ and $\boldsymbol{\phi}_{\epsilon,N,i}$ be an eigenpairs of the Gram matrix of \eqref{transkernel}, that is,
\begin{equation}
\mathbf{K}^{DM}_{\epsilon,M} \boldsymbol{\phi}_{\epsilon,N,i}= \lambda_{\epsilon,N,i}\boldsymbol{\phi}_{\epsilon,N,i}.\label{eigvalDM}
\end{equation}
From \eqref{scalarkernel} and \eqref{eigvalDM}, it is clear that
\[
\sum_{j=1}^N k_{\epsilon,N}(\bx_\ell, \bx_j)\hat{q}_{\epsilon,N}^{1/2}(\bx_j) (\boldsymbol{\phi}_{\epsilon,N,i})_j = \hat{q}_{\epsilon,N}^{1/2}(\bx_\ell) \sum_{j=1}^N k^{DM}_{\epsilon,M}(\bx_\ell, \bx_j) (\boldsymbol{\phi}_{\epsilon,N,i})_j= \lambda_{\epsilon,N,i} \hat{q}_{\epsilon,N}^{1/2}(\bx_\ell) (\boldsymbol{\phi}_{\epsilon,N,i})_\ell,
\]
which suggests that $\boldsymbol{\varphi}_{\epsilon,N,i} = \left(\hat{q}_{\epsilon,N}^{1/2}(\bx_1) (\boldsymbol{\phi}_{\epsilon,N,i})_1,\ldots,\hat{q}_{\epsilon,N}^{1/2}(\bx_N) (\boldsymbol{\phi}_{\epsilon,N,i})_N \right)$ is an eigenvector of the Gram matrix $\mathbf{K}_{\epsilon,N}$ corresponding to the DM kernel \eqref{scalarkernel} with eigenvalue $\lambda_{\epsilon,N,i}$. Thus, the RKHS corresponding to the DM kernel can be characterized as, 
$$\mathcal{H}_{\epsilon,N}=\left\{f = \sum_{i\geq 0}a_i \lambda_{\epsilon,N,i}^{1/2}\psi_{N,\epsilon,i}\right\},$$ where 
$\psi_{N,\epsilon,i}(\bx) = \lambda_{\epsilon,N,i}^{-1}\int_{\mathcal{M}}k_{\epsilon,N}(\bx,\by) \boldsymbol{\varphi}_{\epsilon,N,i}\mu_N(\by)$ is a continuous representation of $\boldsymbol{\varphi}_{\epsilon,N,i}$.

When $\mathcal{M}\subset \mathbb{R}^n$ is a smooth sub-Riemannian manifold, it was shown in \cite{dunson2021spectral,Peoples2025} that the $\lambda_{\epsilon,N,i} \to \lambda_{i}$ as $\epsilon \to 0, N\to \infty$ at an appropriate rate, where $\lambda_i$ denotes the $i$th eigenvalue of the semigroup $e^{-\epsilon \Delta}$ of the heat equation, and $\Delta$ denotes the Laplace-Beltrami operator defined on
manifold $\mathcal{M}$. Likewise, the eigenvector converges to the eigenfunctions of the Laplace-Beltrami operator, $\boldsymbol{\phi}_{\epsilon,N,i} \to \phi_i$, in $L^\infty$-sense over the point cloud data, $X$ \cite{dunson2021spectral,calder2022lipschitz}. 
Since $\hat{q}_{\epsilon,N} \to q^{-1}$ as $\epsilon \to 0, N\to \infty$ pointwisely (see Eq.~(B8) in \cite{coifman2006diffusion}), it is clear that $(\boldsymbol{\varphi}_{\epsilon, N, i})_j \to \varphi_i(\bx_j)$ in $L^\infty$ sense, where $\varphi_i = \phi_i q^{-1/2}:\mathcal{M} \to \mathbb{R}$ is a smooth function. 

The existing result in the literature (see Theorem~16 in \cite{von2008consistency}) shows that one can identify the RKHS for fixed $\epsilon>0$ under the limit of large data $N\to \infty$. To identify the limiting RKHS under $\epsilon \to 0$ and $N\to \infty$, one has to show the convergence of the continuous representation, $\psi_{\epsilon,N,i}$ in $C(\mathcal{M})$. Based on the convergence result in the sense of $L^\infty$ on $X$, our conjecture is that $\psi_{\epsilon,N,i} \to \varphi_i$ as $N\to\infty$ and $\epsilon\to 0$ in $C(\mathcal{M})$. If this conjecture is valid, then  the choice of DM kernel constructs a family of RKHS space $\mathcal{H}_{\epsilon,N}$ that approximates an RKHS space $\mathcal{H}$ corresponding to the kernel,
\begin{eqnarray}
k(\bx,\by) = \sum_{i\geq 0} \lambda_i \phi_i(\bx)q^{-1/2}(\bx)\phi_i(\by) q^{-1/2}(\by).\label{weightedheatkernel}
\end{eqnarray}
Since $\{\phi_i\}$ forms an orthonormal basis of $L^2(\mathcal{M})$, it is clear that $\{\varphi_i\}$ forms an orthonormal basis of $L^2(\mu)$. If $\mu$ is uniformly distributed, then $q$ is constant and the limiting kernel in \eqref{weightedheatkernel} is nothing but the heat kernel, with $\lambda_i = e^{-\xi_i \epsilon}$, where $\xi_i$ denotes the eigenvalue of the (positive definite) Laplace-Beltrami operator $\Delta$ on the Riemannian manifold $\mathcal{M}$.

\section{Numerical results}\label{results}

In this section, we numerically test the proposed DM-KRR on five examples, ranging from smooth to chaotic invariant sets and varying ambient dimensions from 3 to $10^5$.

From these numerical results, We will first show that the proposed DM-KRR outperforms state-of-the-art random feature, neural network, and operator-learning baselines. Second, within our KRR family attained with the proposed dynamic-aware strategy, the diffusion maps (DM) kernel consistently delivers higher accuracy and greater sample efficiency than the Gaussian RBF kernel in \eqref{eqn_rbf}, owing to its ability to encode the intrinsic geometry of the forward invariant set.
In sum, a simple KRR framework with appropriate validation already surpasses modern alternatives, and DM further strengthens performance by adapting to the data geometry.

To evaluate the performance, we use the same metric used for the validation that is defined in Section~\ref{sec:hyperparam} unless otherwise stated. Specifically, we use the RMSE in \eqref{eqn:rmse} for dynamical systems on smooth manifold examples and VPT in \eqref{eq:VPT} for chaotic dynamical systems, except that we now evaluate these metrics on test trajectories of length $T$ rather than on validation trajectories of length $N_V$.

\comment{
\subsection{Evaluation Metrics}

For a trajectory of length $T$, $\{\bx_i\}_{i=1}^T$, suppose the rollout prediction is $\{\hat{\bx}_i\}_{i=1}^T$, where $\hat{\bx}_{i+1}=\bef(\hat{\bx}_i)$ for $i=1,2,\cdots,T-1$, with $\hat{\bx}_1=\bx_1$ and $\bf{f}$ denoting the learned solution operator.  The evaluation metrics are defined as follows,

\begin{enumerate}  
\item \textit{Root Mean Squared Error (RMSE)}: The RMSE is evaluated over the prediction horizon as
\begin{equation}\label{eqn:rmse}
    \text{RMSE}= \sqrt{\frac{1}{Tn}\sum_{i=1}^{T} \left\|\hat{\bx}_i-\bx_{i}\right\|_2^2}.
\end{equation}

\item \textit{Valid Prediction Time (VPT)}:
The VPT is defined as the maximum duration for which the normalized trajectory error remains below a certain error tolerance $\gamma$,
\begin{equation}
    \mathrm{VPT} = \frac{1}{T_{\Lambda}} \max \left\{i|E_i\leq \gamma \text{ for all } i\leq T\right\},
\end{equation}
where $T_\Lambda = 1/\Lambda$ is the unit Lyapunov time, with $\Lambda$ as the maximal Lyapunov exponent of the system.  The normalized error at time step $i$ is
$$
E_i = \sqrt{\frac{1}{n}\sum_{i=1}^n \left(\frac{\hat{x}_{k,i}-x_{k,i}}{\sigma_k}\right)^2 },
$$
where $x_{k,i}$ is the $k$-th component of the state $\bx_i$ and $\sigma_k$ is the standard deviation of the $k$-th component of the true state over the trajectory.
\end{enumerate}
}

\subsection{Dynamics on Torus}

To begin, we first demonstrate the convergence of the KRR methods on the manifold case.  We consider the following system of ODEs that describes the rotation of a rigid body:
\begin{equation}\label{rigid}
\dot z_1 = c_1 z_2 z_3, \quad
\dot z_2 = c_2 z_1 z_3, \quad
\dot z_3 = c_3 z_1 z_2, 
\end{equation}
where $c_1=1/2$, $c_2=-7/8$, and $c_3=3/8$, so that the trajectories with initial conditions on the unit sphere $S^2$ remain on $S^2$ for all time, embedded in $\mathbb{R}^3$.
To evaluate the performance under various ambient dimension, $n$, we map the coordinates $[z_1,z_2,z_3]$ on $S^2$ to coordinates $\bx$ on tori embedded in $\mathbb{R}^n$ for various dimensions $n \in \{3, 7, 15\}$.
However, regardless of the ambient dimension $n$, the forward invariant set of the dynamics is diffeomorphic to three quarters of $S^2$.  See Appendix~\ref{sec:si_case1} for more details on the coordinate transformation.

For training, $100$ initial conditions are sampled on a uniform grid on $S^2$.  For each initial condition, the ODEs are time integrated using the RK4 scheme with $\Delta t = 0.01$ until $t=100$. To strictly enforce the manifold constraint, we normalize the state at every integration step so that its Euclidean norm remains unity.  In addition, to account for the periodic nature of the dynamics, we truncate each path after one period.  In total, there are $1578088$ sample points that are available as training data.
For validation, 12 trajectories of $N_V=301$ steps are generated with the initial conditions uniformly distributed on $S^2$.
For testing, another $32000$ trajectories of $T = 1000$ steps are generated with the initial conditions randomly sampled on $S^2$. Subsequently, we map these datasets to $n$-dimensional torus for training, validation, and testing.

This example uses the KRR models based on DM and RBF using the \textit{skip-connection} form.
The convergence study considers the number of training samples $N=\{1024, 2048, 4096, 8192\}$.
These training samples are randomly selected from the training set in a nested manner, such that the smaller dataset of samples is a subset of the larger dataset.
For each training dataset and each model, we conduct the validation study and determine the hyperparameters that produce the minimum average RMSE over the entire prediction horizon in the validation dataset and compute the mean over the 12 validation trajectories.

\begin{figure}
  \centering
  \includegraphics[width=1\linewidth]{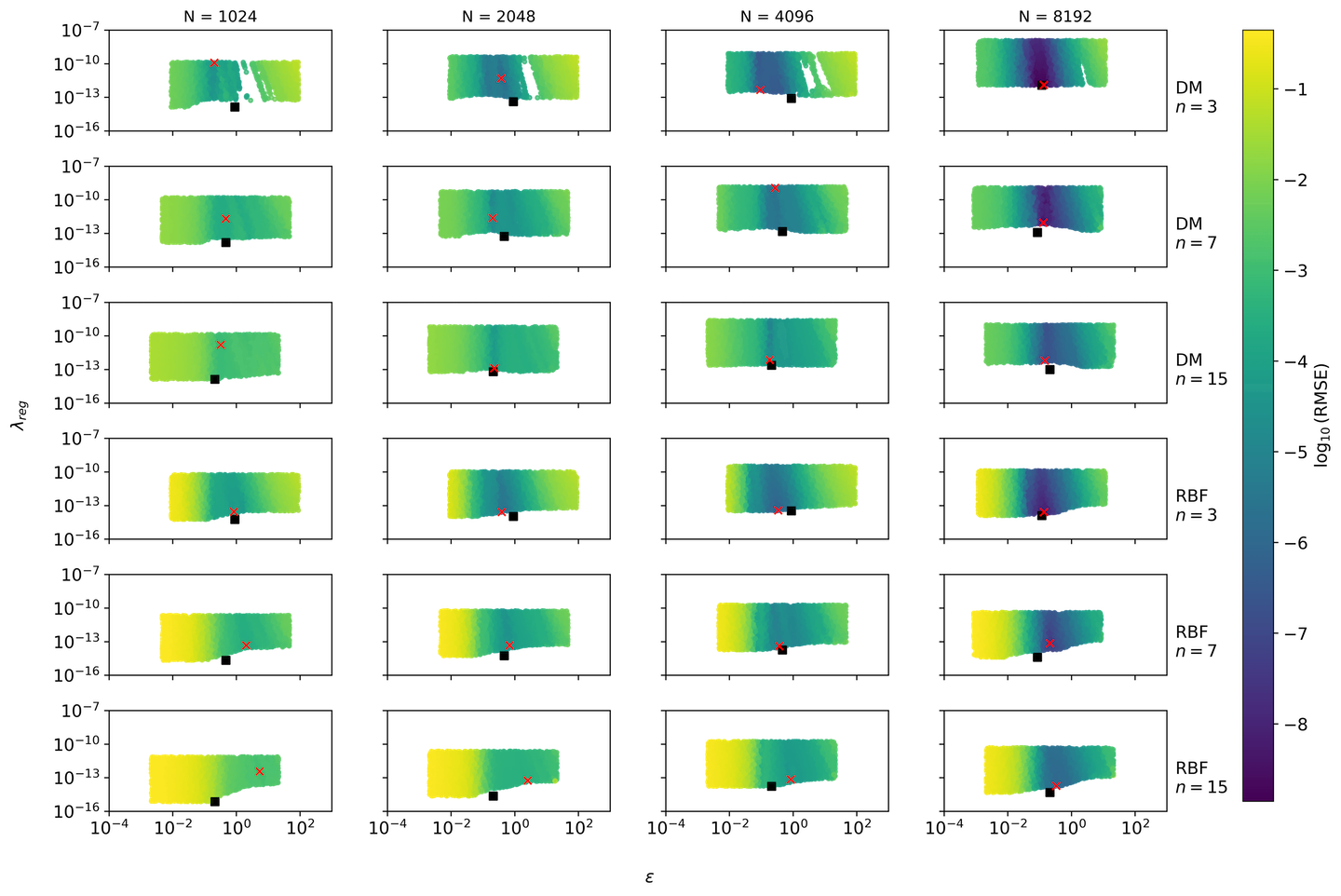}
  \caption{The validation results for torus dynamics. The top three rows show the validation errors of the DM model, and the bottom three rows those of the RBF model. Black squares represent the hyperparameters found using the heuristic approach. Red cross marks represent the best hyperparameter combination found from the validation. The scattered dots represent the trials of the random search; if the RMSE of a trial is too large, the corresponding dot is not shown in the plot.}
  \label{fig:torus-cv}
\end{figure}

The validation results are shown in Fig.~\ref{fig:torus-cv}.  For each number of samples $N$, ambient dimension $n$, and the model, an independent validation study is performed using random search with a maximum of 4096 trials.  The range of random search is determined using the approach provided in Appendix~\ref{sec:si_heur}. Black squares represent the reference hyperparameters found using the heuristic approach.  One thing to notice is that as $N$ increases, reference $\lreg$ increases, which is a typical behavior in the KRR models.  Red cross marks represent the best hyperparameter combination found from the validation.  The scattered dots represent the trials of the random search; if the RMSE of a trial is too large, the corresponding dot is not shown in the plot.  It is noted that as $N$ increases, the black square coincides well with the red cross mark, which demonstrates the accuracy of the reference hyperparameters.  In addition, interestingly this agreement is better for higher ambient dimension as at $N=2048$, they are already aligned well.  


\begin{figure}
  \centering
  \includegraphics[width=1\linewidth]{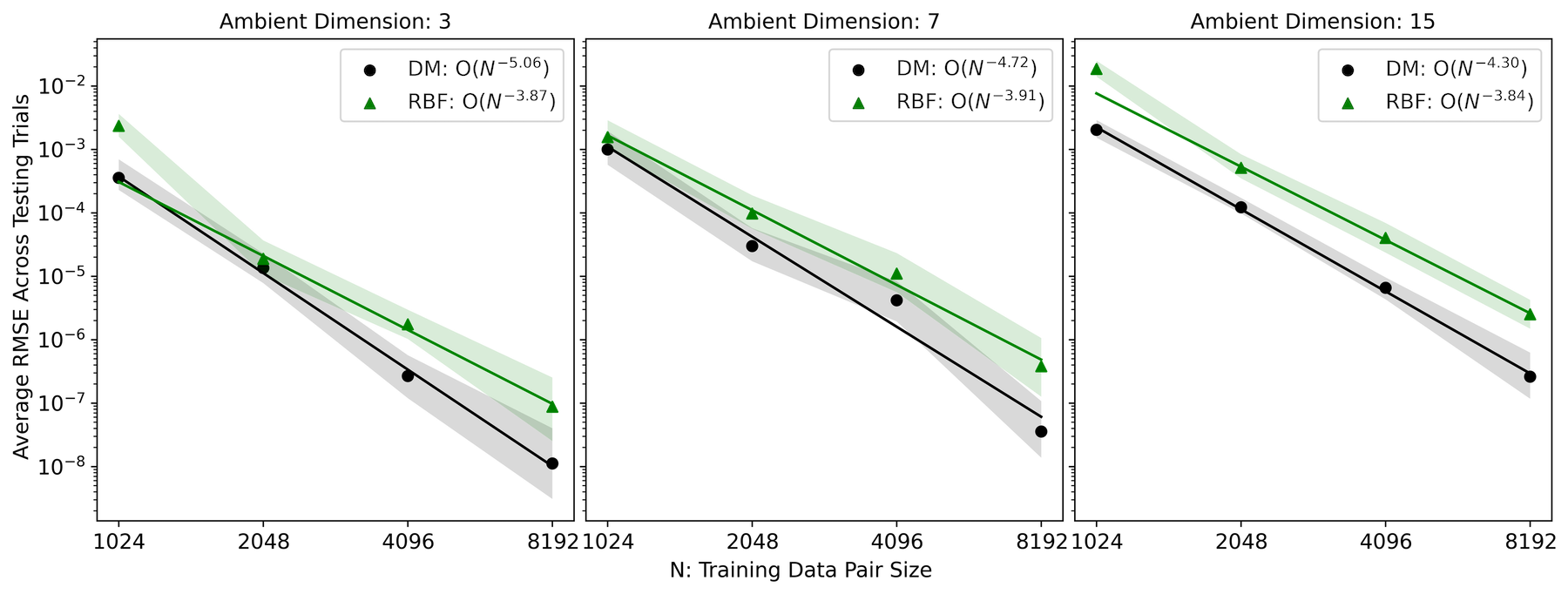}
  \caption{Convergence study on the torus dynamics.  The RMSE's are evaluated on the 32000 test trajectories.  The markers and shades show the mean and min-max bounds of RMSE's, respectively.  Consistently over all the ambient dimensions, DM exhibits a faster convergence rate than the RBF.}
  \label{fig:torus-test-rmse}
\end{figure}

Figure \ref{fig:torus-test-rmse} shows the results of convergence study, with the mean and min-max bounds of RMSE's of the test trajectory.
Consistently over all the ambient dimensions, DM exhibits a faster convergence rate than the RBF.  Furthermore, when the training samples are dense ($N=4096,8192$), the prediction error of DM is approximately one order of magnitude lower than that of RBF.  Lastly, the DM produces RMSEs that follow the fitted convergence rate curves even with relatively scarce training samples ($N=1024$), whereas the RBF's RMSEs are often much larger than the fitted convergence rate curves.

\subsection{Lorenz-63 Dynamics}

Next, we demonstrate the convergence of the KRR methods in the chaotic case.  The standard 3D Lorenz-63 \cite{lorenz1963deterministic} system is considered. The governing equation is:
\begin{equation}\label{lor63}
    \dot{x} = \sigma(y-x), \quad
    \dot{y} = x(\rho-z)-y, \quad
    \dot{z} = xy-\beta z,
\end{equation}
where $\sigma=10$, $\rho=28$, $\beta=8/3$, and $\bx=[x,y,z]$ is used as the states.  The forward invariant set of the system is a strange attractor, of a fractal dimension of approximately 2.06, and the maximal Lyapunov exponent is $\Lambda\approx 0.91$.
To obtain trajectory data, we integrate the Lorenz-63 system using the RK4 scheme with a step size of 0.01s from a randomly sampled initial condition.
To ensure the data is on attractor, we discard the first $4000$ steps, and retain the remaining steps as the dataset.
For training and validation, a single long trajectory of $2\times10^5$ steps is generated.
For testing, we generate another long trajectory of $12.5 \times 10^{5}$ steps, and divide it into 500 non-overlapping segments of length $T = 2500$ steps, where each segment is treated as one test trajectory.
This data generation strategy is consistent with the one reported in \cite{mandal2025learning}.

This example uses the KRR models in \textit{skip-connection} form.
For the convergence analysis, we randomly sample 500 time series of length $N+2N_v$ from the training trajectory to obtain 500 distinct data subsets. In each subset, the first $N$ points are used for training and three overlapping trajectories of length $N_v$ are sampled from the remaining $2N_v$ points for validation. In validation, the VPT criterion of $\gamma=0.3$ is used. In the following subsection, we conduct sensitivity analysis with respect to $N_V$.

\subsubsection{Sensitivity study on validation data size}

To assess the sensitivity of the KRR models with respect to the size of the validation data, we conduct sensitivity study for $N=512$ and $N=1024$.  For $N=512$, 8 validation size were examined, with $N_v$ ranging from $500$ to $1900$ in increments of $200$.  For $N=1024$, 5 cases of validation size are considered; with $N_v$ ranging from $1100$ to $1900$ with the same increment.  Then, for each case of validation data size, 500 models were trained using different subsets of training data.  
For each model 500 VPT's are obtained from the 500 test trajectories, and 500 models produce a total of 250000 VPT's.

\begin{figure}
  \centering
  \includegraphics[width=0.7\linewidth]{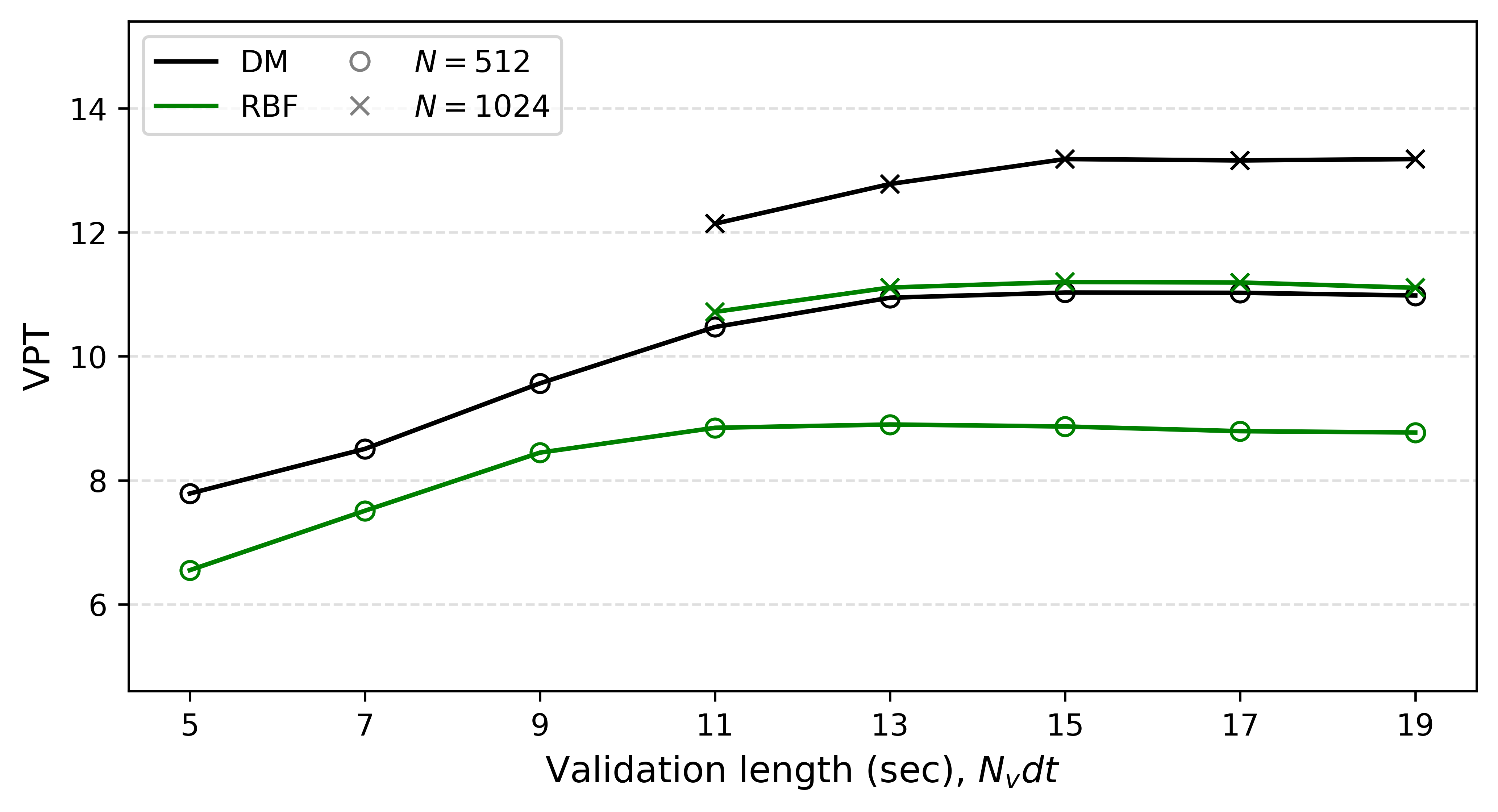}
  \caption{The sensitivity study for Lorenz-63 dynamics.  The mean VPT of DM and RBF for $N=512$ and $1024$ as the validation length increases. The hollow circles correspond to $N=512$, and the cross marks correspond to $N=1024$. In both cases of $N$, DM reaches its maximum VPT at the validation length of 1500. The RBF reaches the maximum at 13 sec for $N=512$ and 15 sec for $N=1024$.  In the horizontal axis, $dt=0.01$ sec.}
  \label{fig:lorenz-sensitivty}
\end{figure}

Figure \ref{fig:lorenz-sensitivty} reports the mean VPT of all 250000 VPT's as $N_v$ increases.  After $N_v\geq 1500$, both DM and RBF do not show significant improvement in mean VPT, which justifies the choice of $N_v=1500$ for the other $N$'s.
Also noteworthy is that, at lower $N_v$, specifically $N_v\leq 700$, the models show a longer VPT than the length of validation trajectory. This is important in applications especially when we don't have long data to validate. Lastly, comparing the two KRR models, DM consistently outperforms RBF for all $N_v$, and the increasing trend in VPT plateaus later than RBF for both $N$; this indicates better representation capability of DM for this chaotic problem.

\subsubsection{Numerical convergence studies}

Based on sensitivity analysis above, we fix $N_v=1500$ for results in this section. Here and in the next section, we do not show the typical validation plot as in Figure~\ref{fig:torus-cv} since the they look somewhat similar to the previous example. 

\begin{figure}
  \centering
  \includegraphics[width=1\linewidth]{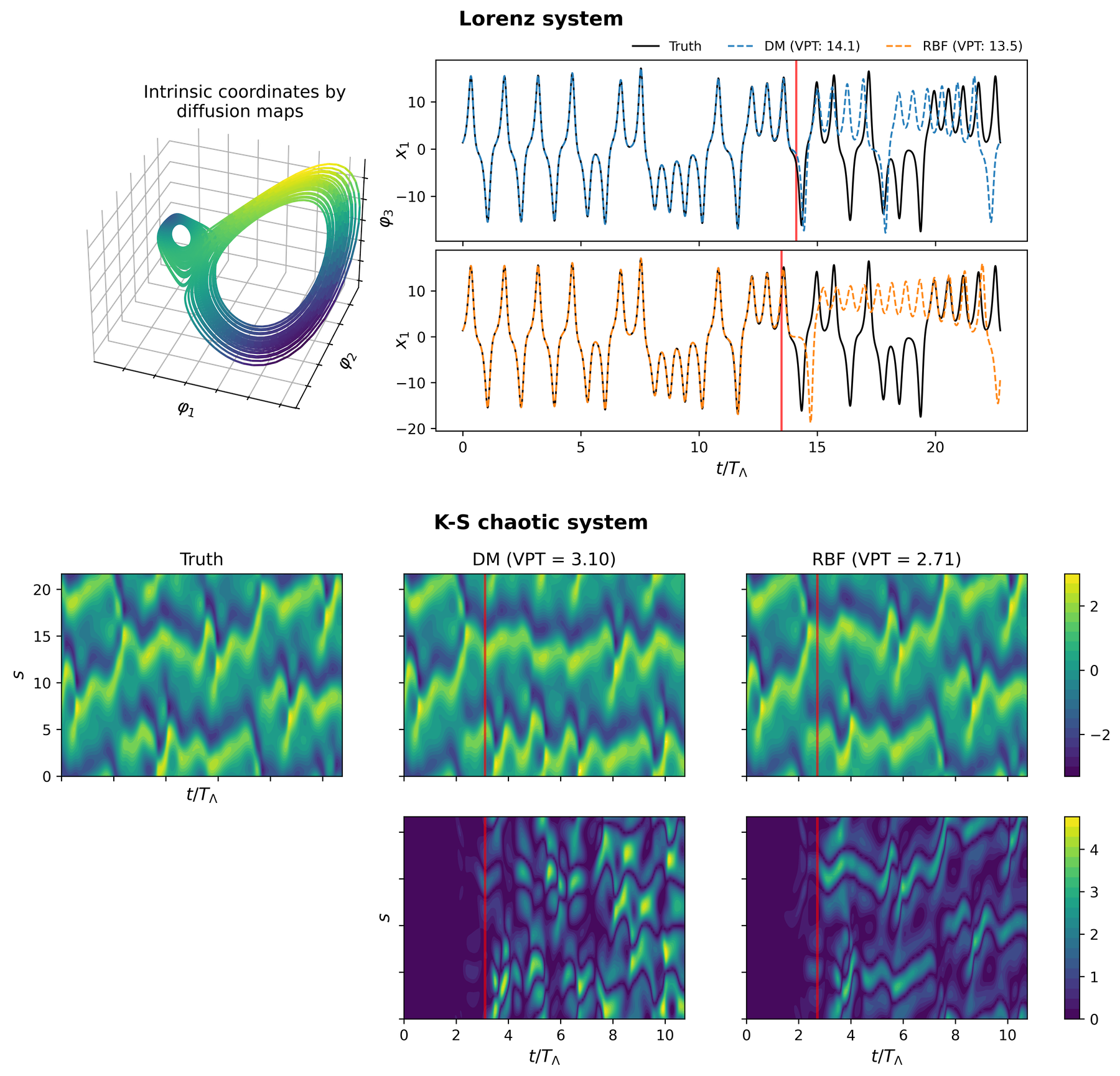}
  \caption{Representative predictions of DM and RBF models for Lorenz and K-S chaotic systems.  Intrinsic coordinates are constructed by the first 3 eigenfunctions of the DM kernel function.  The DM clearly outperforms RBF in terms of VPT in both cases.}
  \label{fig:chaotic-test-pred}
\end{figure}

In the top panel of Fig.~\ref{fig:chaotic-test-pred}, typical model predictions are illustrated for $N=4096$, where DM clearly shows a longer VPT than RBF.
Subsequently, for a more systematic analysis, each of the 500 models is used to compute the VPT's of 500 test trajectories, of which the mean is recorded; this process results in a total of $500$ mean VPT values.
Table~\ref{tab:vpt_lorenz} reports the means and standard deviations of the $500$ mean VPT's for each $N$. As $N$ increases, both KRR models show improved mean VPT, and reduced variance.  Furthermore, across all data sizes, DM consistently outperforms RBF in the mean VPT with lower variance, except $N=512$.  In addition, we highlight that, in the small data size, e.g., $N=1024$, the mean VPT of DM reaches $13.18$ using only $N+2N_v=4024$ data points in total.  Compared to the state-of-the-art random feature model for chaotic dynamics, DeepSkip \cite{mandal2025learning}, which uses 50000 data points, DM achieves 1.16 longer VPT units using over 12 times fewer samples.  

\begin{figure}
  \centering
  \includegraphics[width=1\linewidth]{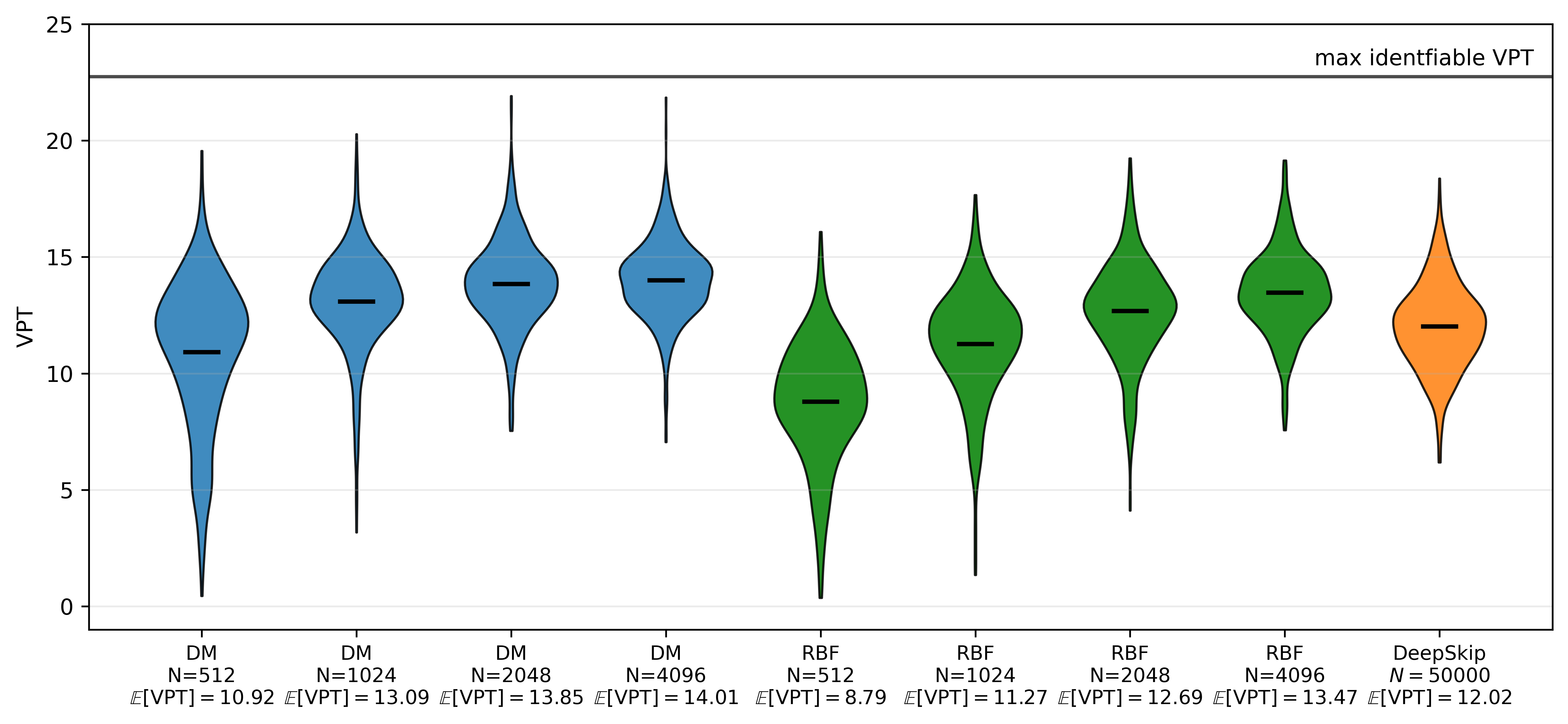}
  \caption{Distributions of VPT of the DM and RBF models for Lorenz-63 dynamics. The max identifiable VPT is the length of test trajectory above which VPT cannot be further evaluated. The rightmost case is the distribution of 500 test VPTs from the best random feature model in \cite{mandal2025learning}.}
  \label{fig:lorenz-test-vpt}
\end{figure}

Figure~\ref{fig:lorenz-test-vpt} represents the distributions of VPT of the DM and RBF models with $N=\{512, 1024, 2048, 4096\}$, compared against DeepSkip \cite{mandal2025learning}.
For a fair comparison, the statistics employed in \cite{mandal2025learning} are reported.
From the 500 models considered in Table~\ref{tab:vpt_lorenz}, we test the models with 500 mutually disjoint test trajectories so that each model is only tested against a single test trajectory.
The violin plot shows the distributions of the 500 VPT's obtained from the 500 KRR models.

The VPT distribution of DeepSkip is also shown in Fig.~\ref{fig:lorenz-test-vpt}, which is also obtained from 500 DeepSkip models \cite{mandal2025learning}. Therefore, it is consistent with the distribution of the other KRR models.  Clearly, even using 50000 training samples, the random feature model still produces a lower mean VPT than most of the KRR models, that are trained and validated using much fewer data samples.

The above results indicate that, with appropriate validation, the KRR models are an accurate approach for long-term predictions in chaotic dynamics, and outperform random feature models, whose training costs are comparable to KRR. Particularly, the DM kernel is more sample-efficient than the conventional RBF kernel, which we attribute to the capability of DM to capture the intrinsic geometry of the dynamics.

\begin{table}[t]
\centering
\caption{Statistics of the valid prediction time (VPT) for Lorenz-63 dynamics.}
\label{tab:vpt_lorenz}
\begin{tabular}{lcccc}
\toprule
\textbf{Model} & \textbf{$N=512$} & \textbf{$N=1024$} & \textbf{$N=2048$} & \textbf{$N=4096$} \\
\midrule
DM  & $\mathbf{11.03 \pm 2.41}$ & $\mathbf{13.18 \pm 1.25}$  & $\mathbf{13.84 \pm 0.59}$ & $\mathbf{14.10 \pm 0.46}$ \\
RBF & $8.87 \pm 1.71$ & $11.20 \pm 1.42$ & $12.72 \pm 0.92$ & $13.47\pm0.61$  \\
\bottomrule
\end{tabular}
\end{table}

\subsection{Kuramoto-Sivashinsky Equation, Chaotic Dynamics}

Subsequently, we demonstrate that the efficacy of the KRR methods remains for chaotic dynamics in higher ambient dimensions.
The KS dynamics is governed by the following partial differential equation (PDE),
$$
u_t+uu_s+u_{ss}+\nu u_{ssss}=0,\quad s\in[0,L],\ t\geq 0.
$$
Here we consider a periodic boundary condition $u(t,0)=u(t,L)$ and a tunable parameter $\nu$.
In this example, we choose $L=22$ and $\nu=1$, and spatially discretize the PDE using a uniform grid of 64 points.
The values of $u$ at the 64 grid points are used as the states, and denoted as $\bx\in\BR^{64}$.
Empirically, the Lyapunov exponent is $\Lambda=0.043$ and the invariant set of chaotic KS dynamics has a dimension of 5.198 (Kaplan-Yorke dimension) \cite{edson2019lyapunov}, much lower than the ambient dimension of $64$.

To generate the training and validation data, we use an initial condition $u(0,s) = \sin(16 \pi s/ L)$ and integrate the system by the Exponential Time Differencing with RK4 (ETDRK4) scheme \cite{kassam2005fourth}.  The step size of 0.01s is used to generate $15.5 \times 10^6$ steps.  The data is downsampled at every 10th step, and the initial $50000$ steps are discarded to remove the transient response; this results in one single trajectory of $15 \times 10^5$ steps.  To generate the test trajectory we use $u(0, s) =  \sin(8 \pi s/ L)$ as the initial condition and follow the same process to generate $12.5 \times 10^5$ sample points, which are divided into $500$ non-overlapping segments of $2500$ steps.

In this problem, we consider DM and RBF models using the \emph{direct} form.  For the convergence study, we follow a similar training and validation procedures in the Lorenz-63 case, but this time using 50 time series of length $N + 2N_v$ from the training trajectory
to obtain 500 distinct data subsets.
We consider $N=\{2048,4096,8192, 16384\}$ and $N_v=2000$, and use VPT criterion of $\gamma=0.5$ for validation. 
\comment{More details of validation are provided in SI \ref{sec:si_case3}.} Lastly, each of the 50 models is tested against 500 test trajectories, of which the mean VPT is recorded; this yields $50$ mean VPT's.

In the bottom panel of Fig.~\ref{fig:chaotic-test-pred}, typical model predictions are illustrated for $N=8192$, where DM shows a longer VPT than RBF.
In Table~\ref{tab:vpt_ks-chaotic}, the mean and standard deviation of the 50 mean VPT's are shown for each $N$ and model.  
Over all $N$'s, DM outperforms RBF in mean VPT, and shows smaller standard deviations except $N=2048$.
Furthermore, the performance gap between DM and RBF widens as $N$ grows, indicating that DM benefits more significantly from increased data availability.


\begin{table}[t]
\centering
\caption{Statistics of the valid prediction time (VPT) for K-S chaotic dynamical system.}
\label{tab:vpt_ks-chaotic}
\begin{tabular}{ccccc}
\toprule
\textbf{Model}  & \textbf{$N=2048$} & \textbf{$N=4096$} & \textbf{$N=8192$} & \textbf{$N=16384$} \\
\midrule
DM  & $\mathbf{0.86 \pm 0.15}$ & $\mathbf{1.72 \pm 0.18} $  & $\mathbf{3.12 \pm 0.21}$ & $\mathbf{4.98 \pm 0.23}$ \\
RBF & $0.79 \pm 0.13$ & $1.49 \pm 0.21$  & $2.72\pm0.26$ &  $4.31 \pm 0.33$ \\
\bottomrule
\end{tabular}
\end{table}

\subsection{Kuramoto-Sivashinsky Equation, Travelling Dynamics}

\begin{figure}
  \centering
  \includegraphics[width=1\linewidth]{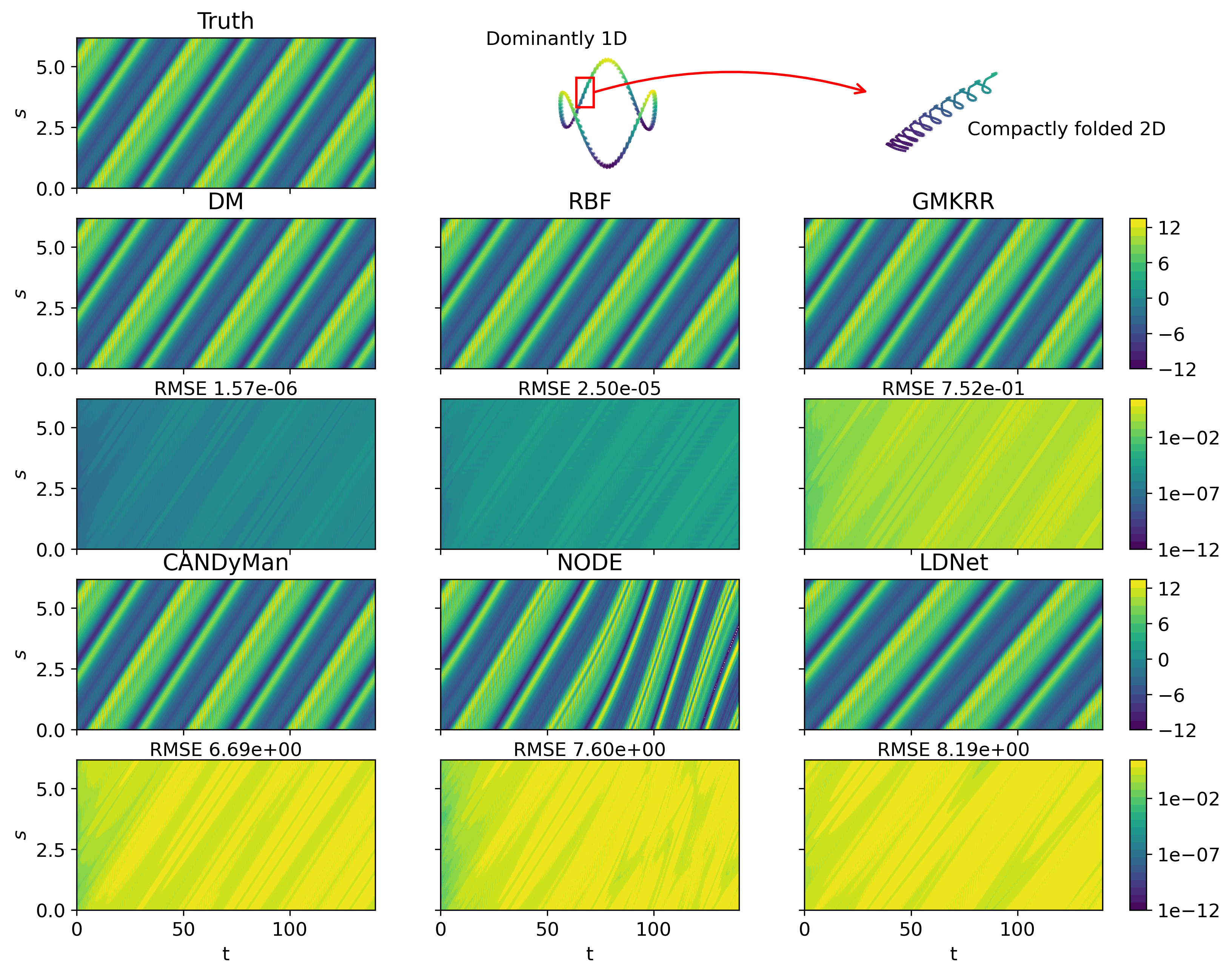}
  \caption{The prediction results for the Kuramoto-Sivashinsky travelling dynamics, comparing the DM and RBF models against four baseline methods.  The DM and RBF models perform drastically better than the baselines over the long timescale, while DM is one order of magnitude better than RBF.}
  \label{fig:ks-travel-test-pred}
\end{figure}

Using the same KS equation, we benchmark the KRR models against other state-of-the-art methods for dynamics having a smooth invariant set.  Specifically, we employ the configuration from \cite{Floryan2022,huang2025learning}, where $L=2\pi$ and $\nu=\frac{4}{87}$ are chosen such that the dynamics exhibit travelling wave behavior.  In this case, the dynamics is intrinsically 2D and the forward invariant set is equivalent to a 2-torus.
Specifically, as shown in the top-left panel of Fig.~\ref{fig:ks-travel-test-pred}, the dynamics has two timescales that differ by a factor of approximately 200. The slow timescale corresponds to the translation of the wave in space, while the fast timescale corresponds to the oscillation of the spatial wave shape.
Furthermore, we also visualize the intrinsic geometry in 3D using the Isomap algorithm \cite{tenenbaum2000global} in the first row of Fig.~\ref{fig:ks-travel-test-pred}.  The middle panel shows the embedding of the full trajectory, that appears to be 1D, but the right panel reveals the highly curled additional dimension that corresponds to the fast timescale.

For data generation, we choose a step size of 0.001s, and the initial condition to be the last step in the training dataset of travelling dynamics in \cite{Floryan2022}.  The simulation is run for $7 \times 10^5$ steps and downsampled at every 10th data points so that the sampling step size is 0.01s. Then, the initial $5 \times 10^4$ steps of transients are discarded, resulting in $2\times 10^4$ steps.  The long period of travelling dynamics is approximately 5000 steps.  We use the first 6000 samples for training and validation, and reserve the remaining 14000 samples for testing.

Here, six models are considered:
\begin{enumerate}
    \item DM: This KRR model uses \textit{skip-connection} form with the DM kernel. Although the training set contains 6000 steps, the data are downsampled at every second point, yielding 3000 training steps used for training. This is to reduce the size of the kernel matrix to achieve efficient validation and test.
    \item RBF: The same as DM, except that the RBF kernel is used.
    \item GMKRR \cite{huang2025learning}: The model is still based on KRR but uses a geometrically constrained operator-valued kernel.  For consistency with the other baseline models, a special data preprocessing step based on prior knowledge (see Appendix \ref{sec:si_case4}) is used for this model, so that the fast and slow timescale dynamics are learned separately, and only the first 100 steps are needed for training.
    \item CANDyMan \cite{Floryan2022}: The model employs a chart-based approach to represent the manifold, and the charts and dynamics are represented using neural networks.  The special data preprocessing is used.
    \item NODE: Using the special data preprocessing, two neural networks are used to represent the vector fields of the fast and slow dynamics, respectively.  No special low-dimensional structure is assumed.
    \item LDNet \cite{Regazzoni2024}: The model has the same setup as NODE, except that the fast dynamic is learned in a 2D latent space.
\end{enumerate}
See Appendix \ref{sec:si_case4} for more details in data preprocessing, model training, and validation.

Figure \ref{fig:ks-travel-test-pred} compares the model predictions.
The baselines NODE and LDNet, that do not account for the exact intrinsic dimension of 2, perform the worst.  In particular, LDNet incorrectly predicts the slow timescale, while NODE fails to predict a constant slow timescale; this is not surprising as the invariant set is not leveraged during prediction.  The CANDyMan performs slightly better than NODE and LDNet, especially in capturing the correct slow timescale, but error accumulates over long-time prediction, likely due to the inaccuracies in the NN approximation.
As GMKRR leverages the prior knowledge like the previous three baselines, its error is one order of magnitude lower than the three, identifying the correct phase.  The enhancement has been attributed to the explicit incorporation of geometrical constraints, including the tangent space of the manifold, in the formulation.  However, the performance of GMKRR depends severely on the accuracy of the estimated tangent space.
Lastly, both the KRR models, DM and RBF, greatly outperform all the previous baselines, where the errors are reduced by 4-5 orders of magnitude.  Furthermore, DM outperforms RBF by one order of magnitude, which is attributed to the capability of DM to implicitly account for the geometrical constraints.

Clearly, the KRR models, especially DM, outperform most baselines even without any prior knowledge or explicit construction of geometrical constraints, and of more practical importance is that the formulation and training of the KRR models are much simpler than the baselines.

\subsection{Pitch-Plunge Flat Plate}

Lastly, we demonstrate the scalability of the KRR models to a real-world problem: an unsteady flow field driven by a pitching and plunging flat plate.  The dynamics is a model problem that appears in many applications, especially the aerodynamics of flying insects and aircraft as well as the fluid dynamics of swimming fish.

A typical snapshot of the flow field is shown in the top-left panel of Fig.~\ref{fig:ppplate-test}, where the contours indicate the vortical strength.  The plate motion generates flow patterns similar to the well-known von K\'arm\'an vortex street.
Specifically, as illustrated in the top row of Fig.~\ref{fig:ppplate-test}, a combined motion of pitching and plunging is prescribed to the flat plate.  Both pitching and plunging follow a harmonic oscillation of frequency $\omega$ and the two oscillations differ by a phase angle $\phi_h$.  The flat plate with prescribed motion results in a moving boundary in the flow field, and thus generating time-varying flow patterns.
Furthermore, the phase angle $\phi_h$ is varied to induce different vortex shedding patterns, involving a rich set of fluid modes.  Prior work \cite{song2025modal} shows that the unsteady flow fields at different phase angles share the same dominant frequencies but consist of different fluid modes.

To generate the datasets, we perform direct numerical simulations of the incompressible Navier–Stokes equations using an immersed-boundary projection method \cite{taira2007immersed, colonius2008fast}.
The simulations are performed on a $599 \times 299$ Cartesian grid, resulting in a state dimension of $n=179101$; details of simulation are provided in Appendix~\ref{sec:si_case5}.
The training data includes $12$ trajectories, using $\phi_h$'s from $0^\circ$ to $330^\circ$ with a step size of $\Delta\phi_h=30^\circ$.  The validation data includes $3$ trajectories using $\phi_h=\{45^\circ,225^\circ,315^\circ\}$.  The test trajectory uses $\phi_h=135^\circ$.
Each trajectory is generated using a step size of 0.01s and then downsampled at every 10th data points; then the first 20s of trajectory is discarded to remove the transient response.  Eventually, $200$ steps are generated for each of the training and validation trajectories, and $801$ steps are generated for the test trajectory.
The long test trajectory, consisting of approximately $20$ periods of flow oscillation, is designed to assess the long-term prediction capability of the models.

This example uses the DM and RBF models using the \emph{direct} form, with comparison to the ResDMD method.  

\subsubsection{Dimension reduction}
Due to the high-dimensionality of the problem, data pre-processing based on PCA is applied to reduce the dimension for the sake of computational efficiency. Specifically, the training data of the $12$ 200-step trajectories are collected as $\bX \in \BR^{n \times 2400}$, $n=179101$.
The mean is $\bar{\bx}=\frac{1}{2400}\bX\mathbf{1}$, where $\mathbf{1}$ is a $2400$-dimension vector of $1$'s, and the mean is subtracted from the training data $\bX'=\bX - \bar{\bx}\mathbf{1}^\top$.
Next, a truncated singular value decomposition is applied, $\bX'\approx \bU_r\vtS_r \mathbf{V}_r^\top$, where $\bU_r\in\BR^{n\times r}, \vtS_r\in \BR^{r \times r}, \bV_r \in\BR^{2400 \times r}$ and $n \ll r$.
Then, a full-dimensional state $\bx$ is reduced as $\tilde{\bx}=\frac{1}{\sigma_1}\bU_r^\top(\bx-\bar\bx)\in\BR^r$, where $\sigma_1$ is the largest singular value and used to normalize the state.

The models are subsequently trained and perform prediction in the $r$-dimensional reduced state space.
We choose $r=81$ for the DM and RBF models, which retains $99.1$\% energy of the data.  More dimensions, $r=162$, are retained for ResDMD, to mitigate its inherent spectral pollution issue. See Appendix~\ref{sec:si_case5} for a brief discussion of the ResDMD method.

Lastly, given a predicted reduced state $\tilde{\bx}$, the full-dimensional flow field is reconstructed as $\hat{\bx} = \sigma_1\bU_r\tilde{\bx} + \bar\bx$. 

\subsubsection{Validation Study}

Due to the multiscale nature of the data set, we consider a different validation metric.
Specifically, we define the following weighted normalized RMSE (WNRMSE) metric.
For $J$ trajectories of length $N_V$ in the reduced space, $\{\tilde{\bx}_{1:T}^{(j)}\}_{j=1}^J$, suppose the rollout prediction is $\{\hat{\bx}_{1:T}^{(j)}\}_{j=1}^J$.  The WNRMSE for $j$ is defined as,
\begin{equation}
    \text{WNRMSE}= \frac{\sqrt{\sum_{j=1}^{J}\sum_{i=1}^{N_V} \hat{w}_i^2\|\hat{\bx}_i^{(j)}-\tilde{\bx}_{i}^{(j)}\|_2^2}}{\sqrt{\sum_{j=1}^{J}\sum_{i=1}^{N_V} \|\tilde{\bx}_{i}^{(j)}\|_2^2}}, \quad \hat{w}_i = \frac{w_i}{\sqrt{\sum_{i=1}^Tw_i^2}}
\end{equation}
where the weight $w_i = \exp(t_i-t_T)$ gives more penalty near the final time, and promotes the hyperparameters for more accurate predictions in long time horizon. 

\begin{figure}
  \centering
  \includegraphics[width=0.7\linewidth]{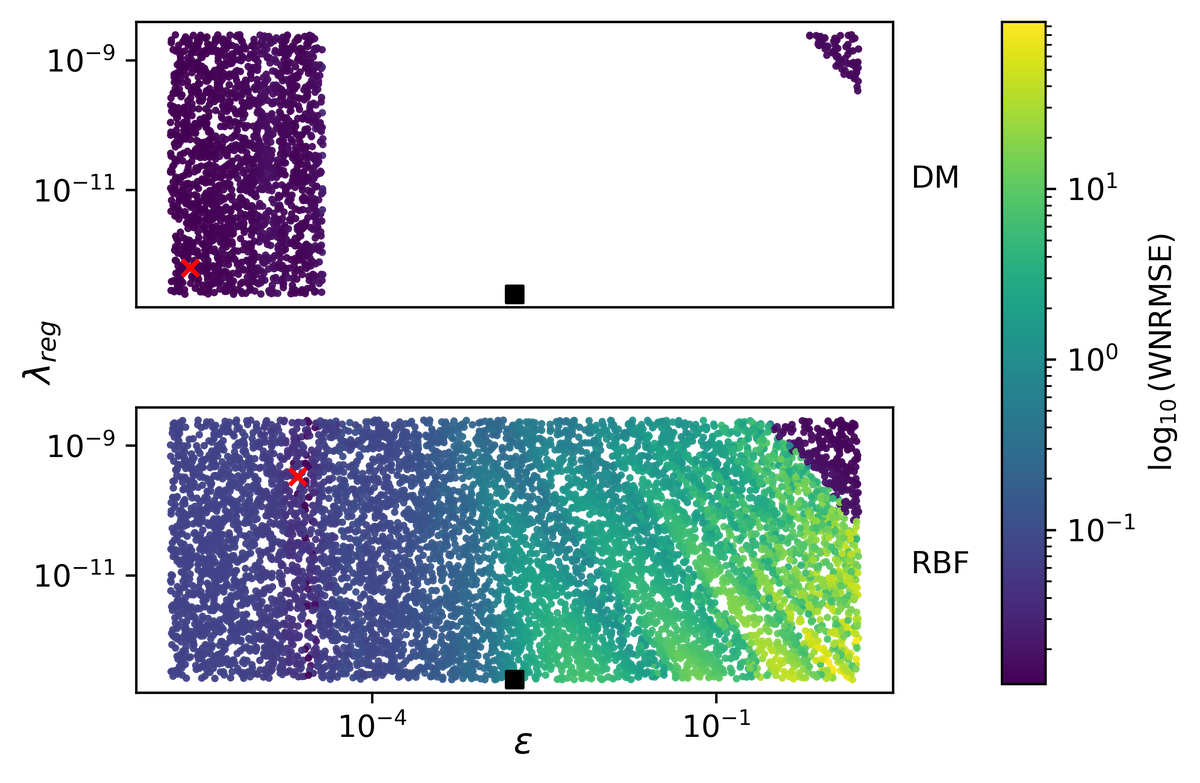}
  \caption{The validation results for the pitch-plunge flat plate for the DM and RBF models.  Black squares represent the hyperparameters found using the heuristic approach. Red cross marks represent the best hyperparameter combination found from the validation. The scattered dots represent the trials of the random search; if the RMSE of a trial is too large, the corresponding dot is not shown in the plot.}
  \label{fig:ppplate-cv}
\end{figure}

For the KKR models, we employ a broader search range than other cases, $\Delta \epsilon=10^{3}$, for the random search in validation. This is because the dataset is relatively sparse, and the heuristic estimates of lengthscales may be less accurate.  To account for the broader search domain, the number of search trials is doubled to be 8192.
The validation results are shown in Fig.~\ref{fig:ppplate-cv}.

For DM, the predictions diverge for a substantial portion of the search space, leading to the large empty region visible in Fig.~\ref{fig:ppplate-cv}; RBF also exhibits high errors in the same region.  There is also a region on the top-right side that seems appropriate for both models, but the validation study did not select hyperparameters from this region for DM or RBF.
Specifically, for DM, hyperparameters from this region produce predictions that decay over long horizons.  For RBF, hyperparameters in this region do provide more accurate prediction than the one presented below, however, they give decaying response in the long horizon and is still worse than the corresponding DM model.  The validation results indicate that the lack of data, i.e., longer prediction horizon, poses challenge to validation for RBF, but less so for DM. In the next section, we show prediction results based on the best hyperparameters selected in Figure~\ref{fig:ppplate-cv}.

\subsubsection{Comparisons of the predictions}

\begin{figure}[htbp]
  \centering
  \includegraphics[width=1\linewidth]{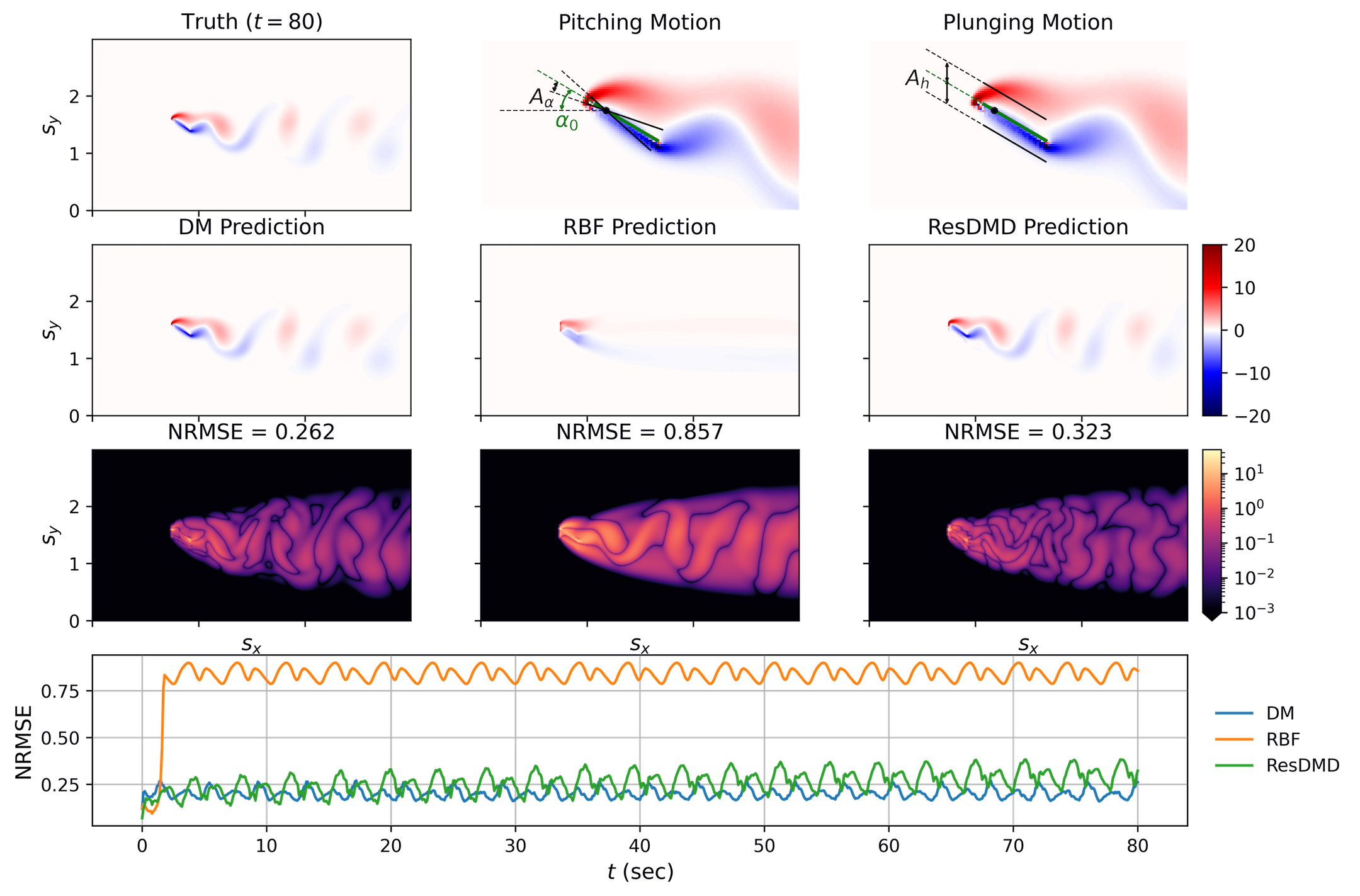}
  \caption{The prediction results for the pitching and plunging plate at $t=80$, comparing the DM, RBF, and ResDMD models.  The RBF model fails likely due to the data sparsity, while DM outperforms ResDMD in accuracy and maintains the same level of accuracy over a long prediction horizon.}
  \label{fig:ppplate-test}
\end{figure}

The typical snapshots of the model predictions and NRMSE's are shown in the middle two rows of Fig.~\ref{fig:ppplate-test} for the test trajectory.  The NRMSE at step $i$ shown in the third row is defined as 
$
    \text{NRMSE} = {\| \hat{\bx}_i -  \bx_i\|_2}/{\| \bx_i \|_2}
$, where the norm is over the space.
Clearly, the RBF fails to capture the system dynamics and the vortex shedding phenomenon is completely missing.  In contrast, DM and ResDMD provide relatively accurate predictions, where DM has the lower error.
The last panel of Fig.~\ref{fig:ppplate-test} shows the temporal evolution of the error over the entire test prediction horizon.
Again RBF deviates substantially from the truth even in the beginning of the horizon.
There is a noticeable growth of error in the ResDMD model; this is due to the inaccurate eigenvalues identified by the model.
In contrast, DM not only achieves the lowest overall error among the three models, but also consistently maintains the error level over the horizon, indicating that it is respecting the geometrical constraint over a long-term prediction of $20$ cycles. Comparisons of results at shorter prediction times are shown in Appendix~\ref{sec:si_case5}.

\section{Conclusion and Discussion}\label{summary}

This work demonstrates that kernel ridge regression (KRR) equipped with a scalar-valued, data-driven kernel provides a simple yet highly effective framework for learning solution operators of dynamical systems across a wide range of regimes. Through systematic evaluation on smooth manifolds, chaotic attractors, and high-dimensional spatiotemporal flows, we showed that when combined with an appropriate, dynamics-aware validation procedure, KRR reliably outperforms state-of-the-art random feature, neural network and operator-learning approaches in both long-term predictive accuracy and sample efficiency. These findings highlight an important principle: much of the difficulty associated with data-driven dynamics learning arises not only from insufficient model complexity, but also from mismatched model selection criteria. With appropriate validation, even a simple scalar-kernel KRR formulation is competitive with or superior to modern, architecture-heavy baselines.

Within this framework, the diffusion maps (DM) kernel offers additional advantages by encoding the intrinsic geometry of the forward invariant set. Across all experiments, DM consistently achieved lower error and required fewer samples than the conventional Gaussian RBF kernel. This suggests that geometry-adaptive kernels can provide a substantial advantage even without explicit manifold reconstruction or tangent-space estimation. The DM kernel's ability to approximate the heat kernel on manifold further aligns the learned operator with the underlying geometric structure, offering both empirical robustness and theoretical motivation.

Several directions for future work naturally follow.
On the theoretical front, there are several open problems. First, the characterization of the limiting Reproducing Kernel Hilbert Space that is being approximated under finite $N, \epsilon$ is an important question. Second, it remains important to establish convergence guarantees for the learned solution operators, especially for chaotic systems where invariant sets may be fractal and data correlations violate standard i.i.d.~assumptions. 
From a computational standpoint, the cubic scaling of exact KRR motivates investigating the potential of using fast solvers to enable larger datasets or real-time deployment, adopting recently developed methods such as randomized sketching techniques
\cite{zhao2024adaptive} and sparse Cholesky decomposition \cite{schafer2021}.
In addition, it is known that the fixed bandwidth implementation of the DM, as used in this paper, can struggle when the underlying manifold has a non-uniform density that has areas of especially low density. One possible solution to this problem is to use the variable bandwidth DM \cite{berry2016variable} to learn the intrinsic geometry of the manifold more accurately and potentially improve the predictive accuracy.
Practically, another consideration is data noise, which can be detrimental to the predictive capability of the proposed framework; a possible solution is to couple KRR to a Kalman filter framework for data denoising.
Finally, integrating adaptive sampling or multi-trajectory experimental design could further reduce data requirements in high-dimensional PDE settings.

\textbf{Data availability}
The pitching and plunging flat plate data are available at \url{https://doi.org/10.5281/zenodo.17932009}. The rest of data can be generated by the code at the GitHub repository \url{https://github.com/JWS625/diffusion_maps}.

\textbf{Code availability}
The code for reproducing the results presented in this paper is publicly available on GitHub at \url{https://github.com/JWS625/diffusion_maps}. The code is written in Python and utilizes CuPy for the GPU implementation of KRR models for both DM and RBF kernels.

\textbf{Acknowledgments.} 
The work was partially supported by the NSF Grant DMS-2505605.
The research of J.H. was partially supported by the Office of Naval Research (ONR) grant N000142212193.  The research of D.H. was partially supported by the NSF Grant CMMI-2340266.

\textbf{Author contributions.}
J.H. and D.H. supervised the study. J.H. contributed to the initial concept. D.H. and J.S. carried out model development, implementation, training, analysis, and visualization. All authors contributed to writing the manuscript.

\textbf{Competing interests.} 
The authors declare no competing interests.

\bibliographystyle{plain}
\bibliography{refs}

\appendix

\section{Heuristic estimation of the kernel lengthscale}\label{sec:si_heur}

For the efficient search of the hyperparameters, it is necessary to start with a good initial guess, especially for the lengthscale.  Here we take a strategy that is modified from \cite{huang2025learning} to estimate the lengthscale.  The core idea is to leverage a kernel-based intrinsic dimension estimation algorithm, which produces a bandwidth parameter as a byproduct; we then estimate lengthscale for KRR kernels based on the bandwidth.

Specifically, we use the intrinsic dimension estimation algorithm provided in \cite{coifman2008graph} using a Gaussian RBF kernel.  For numerical robustness, we choose ${\rho}(\bx,\bx';\eta)=\exp\left(-\norm{\bx-\bx'}^2/(L^2\eta)\right)$, where $L$ is the maximum of pair-wise L2 distances of the dataset and $\eta$ is the bandwidth parameter.

Consider a dataset of points $\{\bx_i\}_{i=1}^N$ that may lie on an unknown manifold of intrinsic dimension $d$. Define the sum
$$
S(\eta) = \frac{1}{N^2}\sum_{i,j=1}^N \rho(\bx_i,\bx_j;\eta).
$$
One can verify that $\lim_{\eta\rightarrow 0}S(\eta)=1/N$ and $\lim_{\eta\rightarrow\infty}S(\eta)=1$.  Furthermore, \cite{coifman2008graph} shows that $S(\eta)\propto \eta^{d/2}$ in a range of $\eta$, which leads to the following estimation of the intrinsic dimension.  First define
$$
\tilde{d}(\eta) = \frac{2\mathrm{d}\log(S(\eta))}{\mathrm{d}\log(\eta)},
$$
then the estimated intrinsic dimension is $d^*=\tilde{d}(\eta^*)$, where
$$
\eta^* = \arg\max_\eta \tilde{d}(\eta).
$$
In the numerical implementation of the above intrinsic dimension estimation method, the derivative $\tilde{d}(\eta)$ can be evaluated analytically, since $\tilde{d}(\eta)=\left.\frac{\eta}{S}\right/\frac{\mathrm{d}S}{\mathrm{d}\eta}$, or simply by finite difference.

Finally, while the above bandwidth parameter $\eta^*$ is developed for intrinsic dimension estimation, we found empirically that a good initial guess for the lengthscale parameter $\epsilon^*$ in the RBF and DM kernels can be computed based on $\eta^*$.  For manifold problems, we choose
$$
\epsilon^* = \frac{5}{2n}(L^2\eta^*)^{1/d^*}.
$$
For chaotic problems, while the forward invariant set is not smooth, we still use the above algorithm and choose
$$
\epsilon^* = 250 (L^2\eta^*).
$$

\section{Dynamics on Torus}\label{sec:si_case1}

\begin{figure}
  \centering
  \includegraphics[width=1\linewidth]{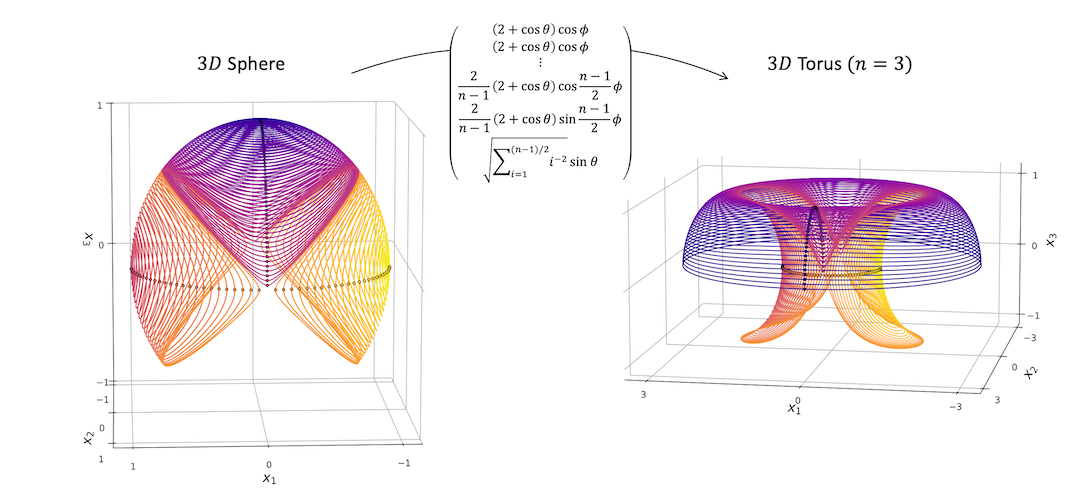}
  \caption{Visualization of the transformation from a unit sphere to a torus.}
  \label{fig:sphere-torus-vis}
\end{figure}

\paragraph{Coordinate transformation.}
Given a point $[z_1,z_2,z_3]$ on a unit sphere $S^2$, we first compute its spherical coordinate representation as
$$
\theta = \arctan \frac{z_2}{z_1},\quad \phi = \arccos z_3.
$$
We then use $(\theta, \phi)$ to map onto a general $n$-dimensional ambient space by the following formula:
\begin{equation}\label{torus-formula}
(\theta, \phi) \mapsto
\begin{pmatrix}
  (2+\cos \theta) \cos \phi \\
  (2 + \cos \theta) \sin \phi \\
  \vdots\\
  \frac{2}{n-1}(2+\cos \theta) \cos\frac{n-1}{2}\phi \\
  \frac{2}{n-1}(2+\cos \theta) \sin\frac{n-1}{2}\phi \\
  \sqrt{\sum_{i=1}^{(n-1)/2} i^{-2}}\sin \theta
\end{pmatrix},
\end{equation}
where $n>1$ is an odd integer.  An example of the transformation when $n=3$ is shown in Fig.~\ref{fig:sphere-torus-vis}.
Note that, we only consider the range $(\theta, \phi)=[0,\pi/2]\times[0, \pi]$ to ensure that the transformed vector field on the torus is continuous.

\section{Kuramoto-Sivashinsky Equation, Travelling Dynamics}\label{sec:si_case4}

\paragraph{Data preprocessing.}
As explained in the main text, to ensure the best prediction performance, the baseline models, GMKRR, CANDyMan, NODE, and LDNet, all use a special data preprocessing step, which is explained below.

Denote the data sample at time $t$ as $\bu(t)\in\BR^{64}$.  The preprocessing starts with a discrete Fourier transform to decompose the data $\bu(t)=\mathrm{Re}(\hat{\bu}(t)\exp(i\phi(t)))$ into a standing wave $\hat{\bu}(t)$ and its phase shift $\phi(t)$.  The standing wave and phase dynamics capture the fast and slow timescales, respectively, and both have an intrinsic dimension of 1.  After the preprocessing step, the four models only needed the first 100 steps to train for the standing wave dynamics $\dot{\hat{\bu}}=f_{NN}(\hat{\bu})$ as well as the change in phase per time step $\Delta{\phi}=g_{NN}(\hat{\bu})$.

\begin{figure}
  \centering
  \includegraphics[width=0.7\linewidth]{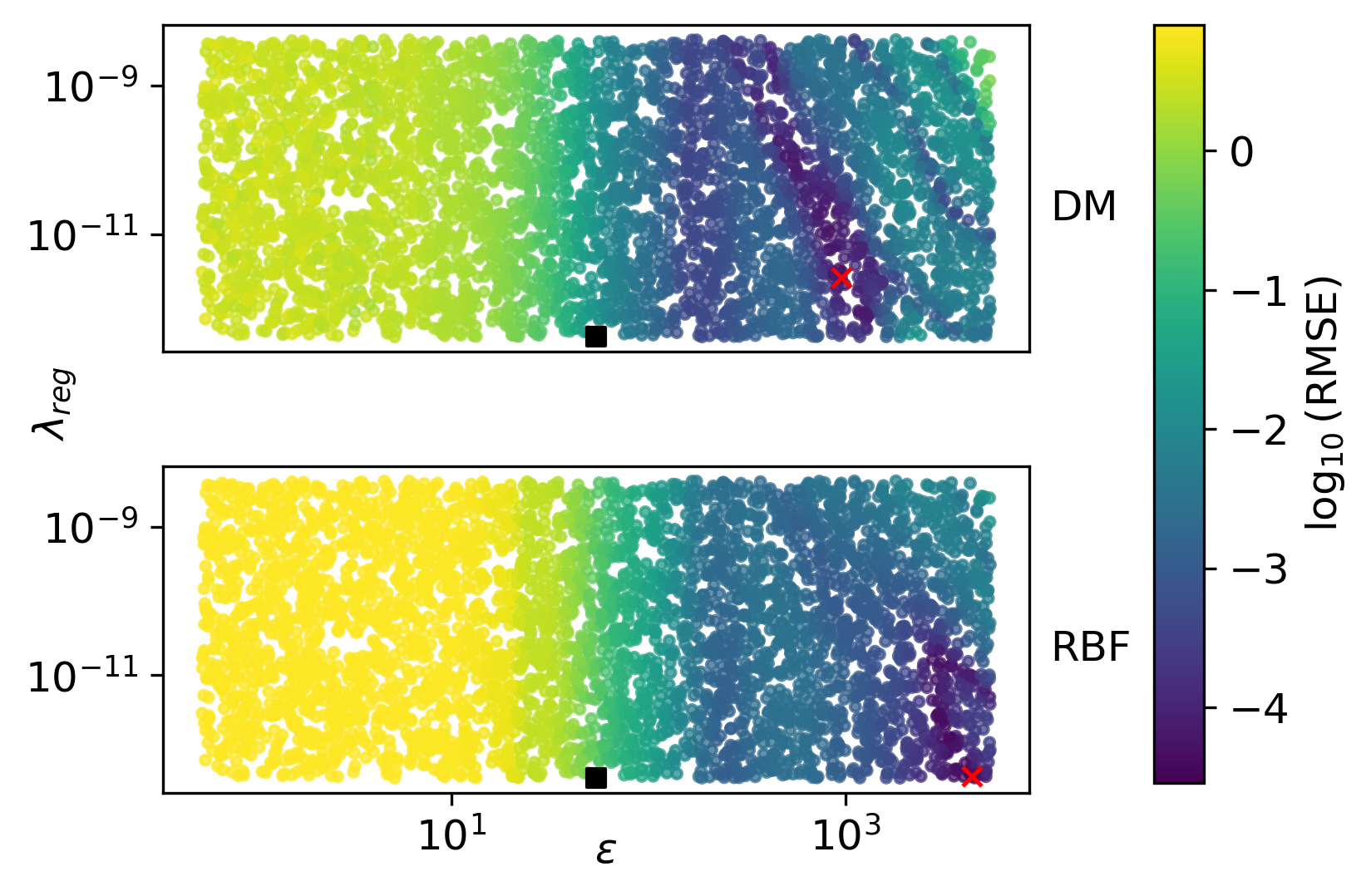}
  \caption{The validation results of the DM and RBF models for KS traveling dynamics. Black squares represent the hyperparameters found using the heuristic approach. Red cross marks represent the best hyperparameter combination found from the validation. The scattered dots represent the trials of the random search; if the RMSE of a trial is too large, the corresponding dot is not shown in the plot.}
  \label{fig:ks-traveling-cv}
\end{figure}

\paragraph{Validation for the KRR models.}
The validation results are shown in Fig.~\ref{fig:ks-traveling-cv}, where the hyperparameters are determined by minimizing the RMSE over the validation trajectories.  For this system, because of the fast and slow scale dynamics, it seems challenging to find a good $\epsilon^*$ from the intrinsic dimension estimation; the fast dynamics evolve tightly over time, leading to numerical difficulty in the algorithm.  In fact, the algorithm identifies the intrinsic dimension as $3$ which is supposed to be $1$, and this is the indication of that the algorithm cannot extract $1D$ information from the compactly folded data.  Nevertheless, both models are able to find the optimal hyperparameters within the $\Delta\epsilon=10^{-2}$ bounds.

\paragraph{Hyperparameters of the baseline models.}
The hyperparameters for CANDyMan are the same as the original paper \cite{Floryan2022}, where its atlas-of-charts approach is applied to $f_{NN}$ and $g_{NN}$ is a vanilla fully-connected NN (for each chart).  For the NODE, $f_{NN}$ consists of 64 inputs and 64 outputs, with two hidden layers of 128 neurons each and $\tanh$ activation function, and the rest is the same as the 1D case, except that 2500 iterations were used in training; $g_{NN}$ consists of one hidden layer of 32 neurons and $\tanh$ activation function with 1000 iterations, and is trained by Adam optimizer with learning rate $10^{-3}$.
The LDNet is the same as NODE, except the $f_{NN}$ is replaced by the latent dynamics.
The dynamics network has 64 inputs and two outputs (latent variables), with two hidden layers of 8 neurons each and $\tanh$ activation function; the reconstruction network has three inputs (latent variables and $x$-coordinate) and one output (the value of one state), with two hidden layers of 8 neurons each and $\tanh$ activation function.
The rest is the same as the 1D case, except that 6000 iterations were used in training.  In this case LDNet use the same dataset as other models.  Note that LDNet would fail to produce reasonable predictions if only one latent variable (i.e., the intrinsic dimension) is used.

\section{Pitch-Plunge Flat Plate}\label{sec:si_case5}

\begin{figure}
  \centering
  \includegraphics[width=0.7\linewidth]{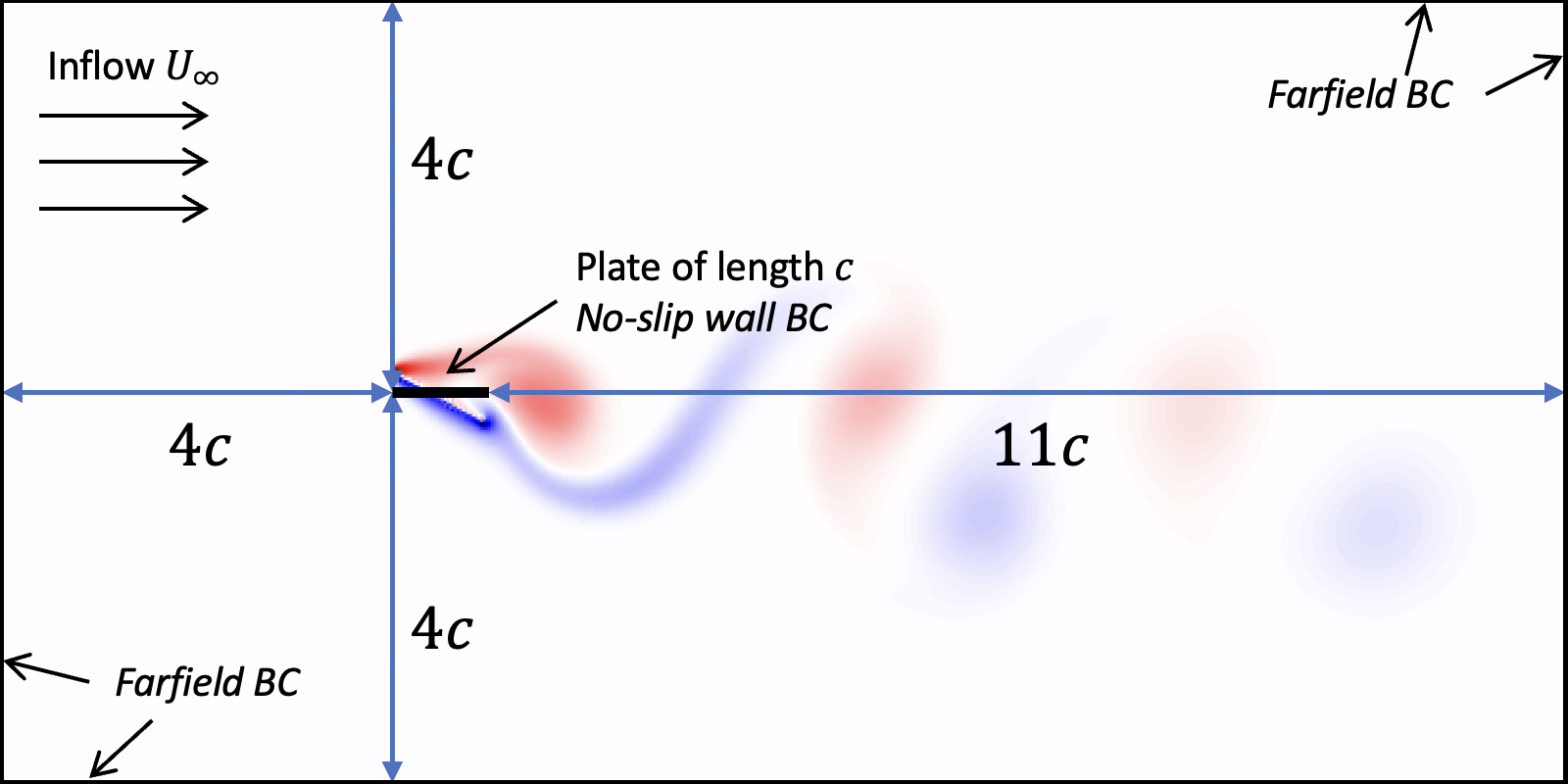}
  \caption{The computational domain of pitch-plunge flat plate.} 
  \label{fig:ppplate-domain}
\end{figure}

\paragraph{Direct numerical simulation (DNS).}
The DNS is based on the open-source code from \cite{taira2007immersed, colonius2008fast}.
The flat plate has a unit length, denoted by $c$.
The computational domain is shown in Fig.~\ref{fig:ppplate-domain}, which is $8c$ high and $16c$ wide.  The farfield boundary conditions are prescribed to all sides of the domain, and no-slip wall boundary condition is prescribed to the flat plate.
The domain is discretized evenly into a $599\times 299$ Cartesian grid.
The inflow velocity is chosen such that the Reynolds number is $100$, with $c$ as the reference length.

The pitching is with respect to the quarter chord of the plate, and the pitching angle is prescribed as $\alpha(t) = A_{\alpha} \sin(\omega t) + \alpha_0$; the plunging is prescribed as $h(t) = A_h \sin(\omega t + \phi_h)$.
In this study, we fix $\alpha_0=30^\circ$, $A_\alpha=5.730^\circ$, and $A_h=0.2c$.  We set the frequency as $\omega\approx 1.602 \mathrm{rad/s}$, which is the vortex shedding frequency when the flat plate is static with $\alpha_0=30^\circ$.

\paragraph{Data visualization.}

\begin{figure}[t]
    \centering
    \begin{subfigure}[b]{0.4\textwidth}
        \centering
        \includegraphics[width=\textwidth]{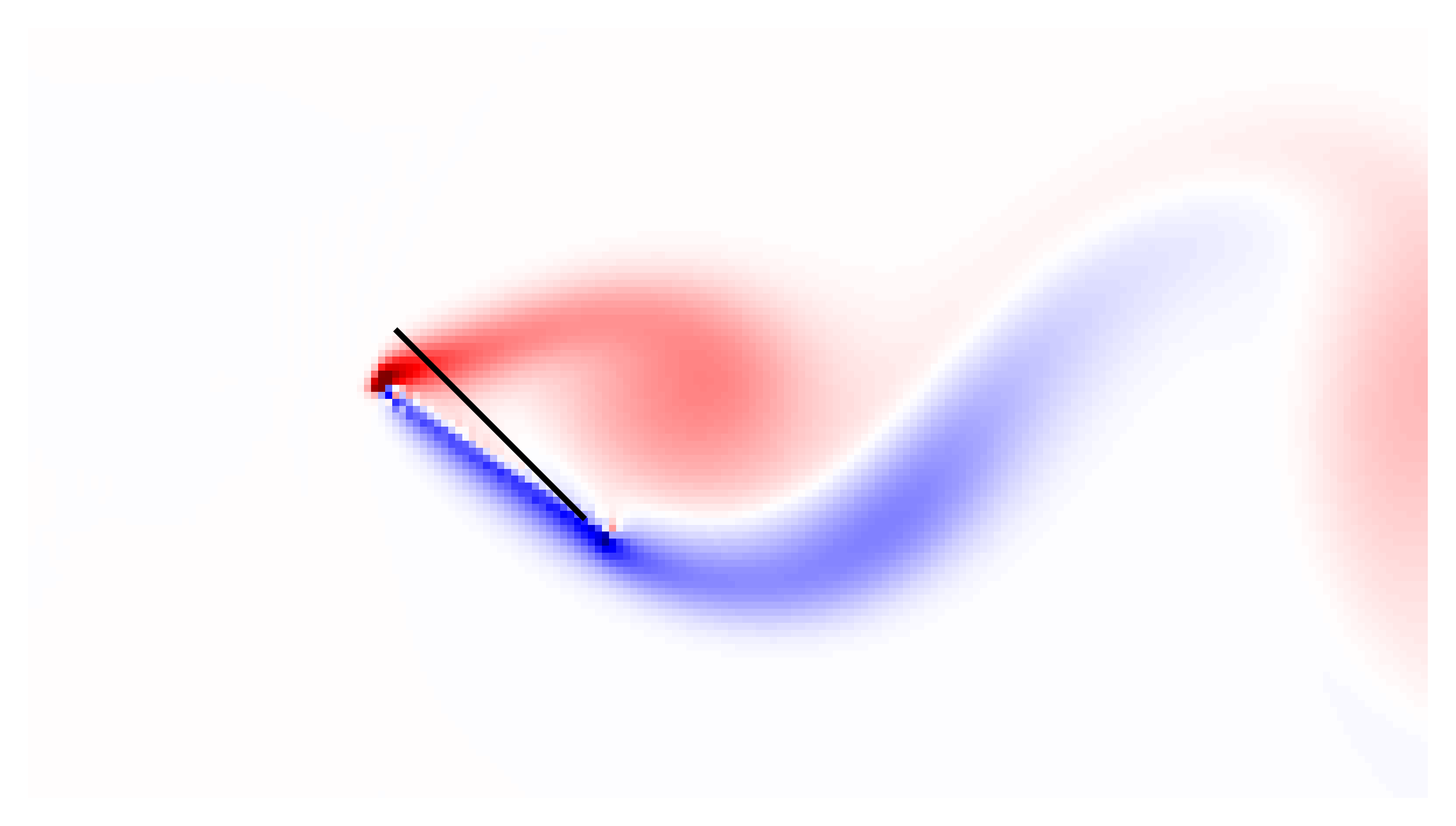}
        \caption{$t=0$}
        \label{fig:ppplate-1s}
    \end{subfigure}
    \begin{subfigure}[b]{0.4\textwidth}
        \centering
        \includegraphics[width=\textwidth]{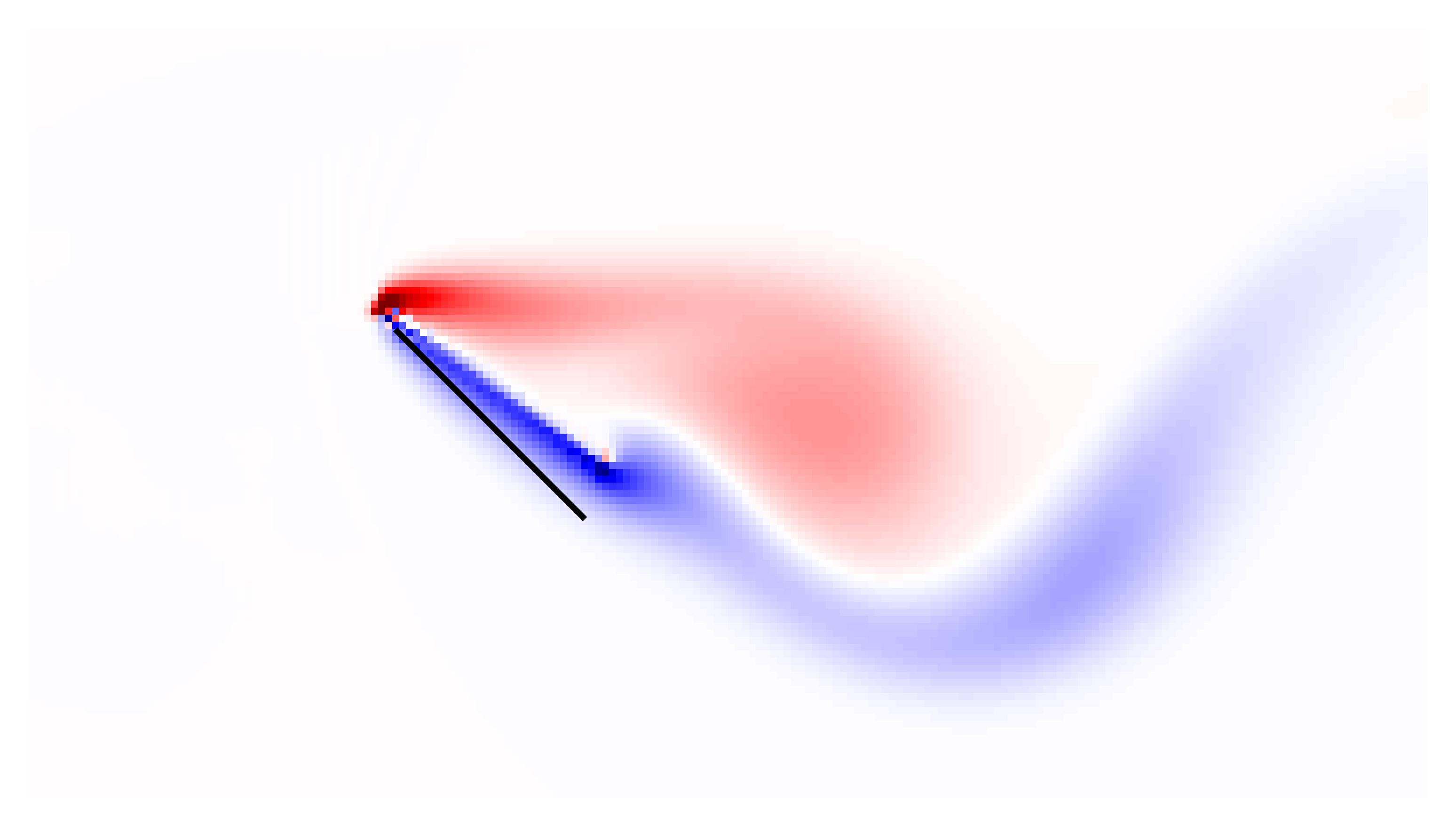}
        \caption{$t=1$}
        \label{fig:ppplate-2s}
    \end{subfigure}
    \begin{subfigure}[b]{0.4\textwidth}
        \centering
        \includegraphics[width=\textwidth]{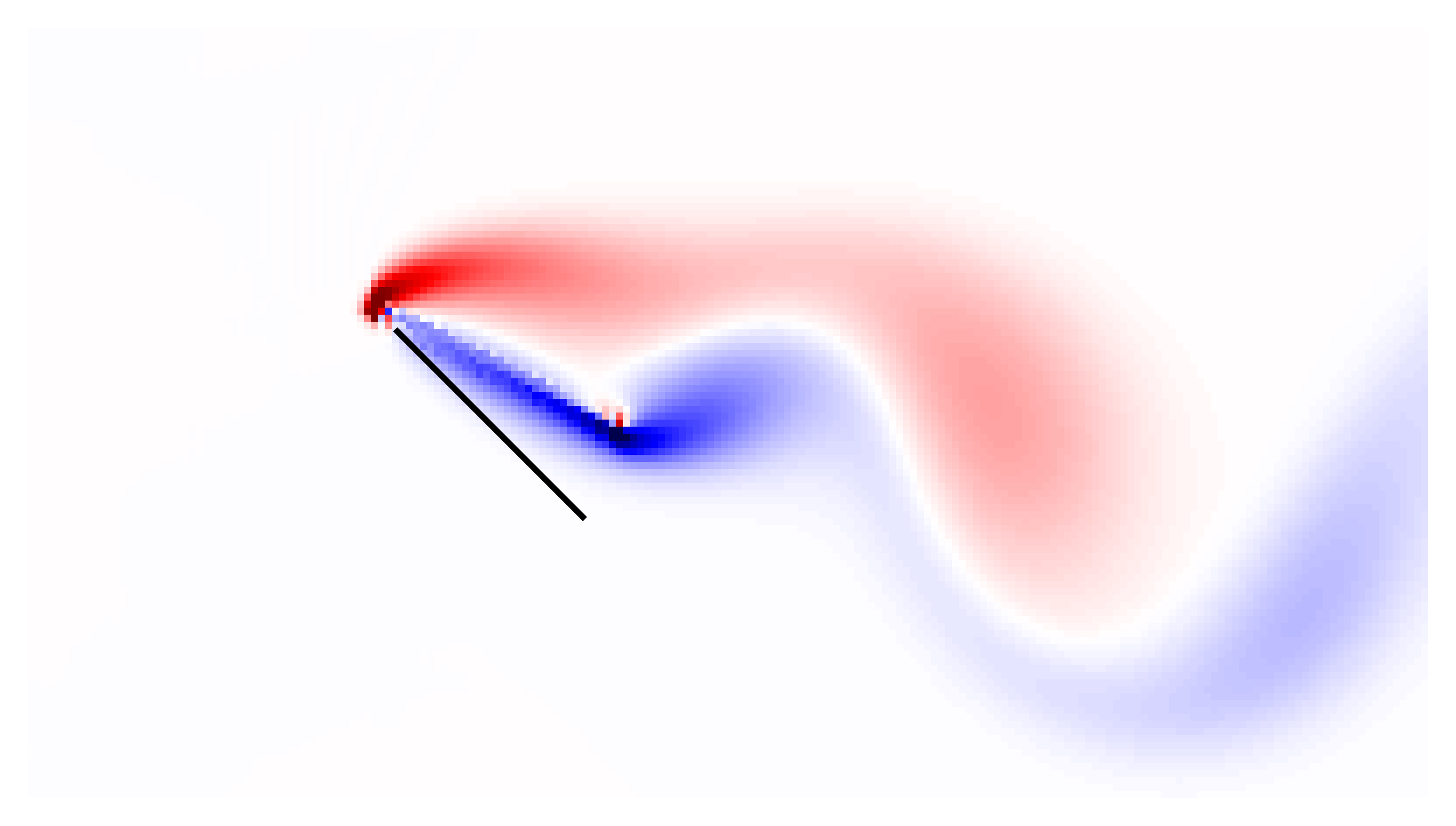}
        \caption{$t=2$}
        \label{fig:ppplate-3s}
    \end{subfigure}
    \begin{subfigure}[b]{0.4\textwidth}
        \centering
        \includegraphics[width=\textwidth]{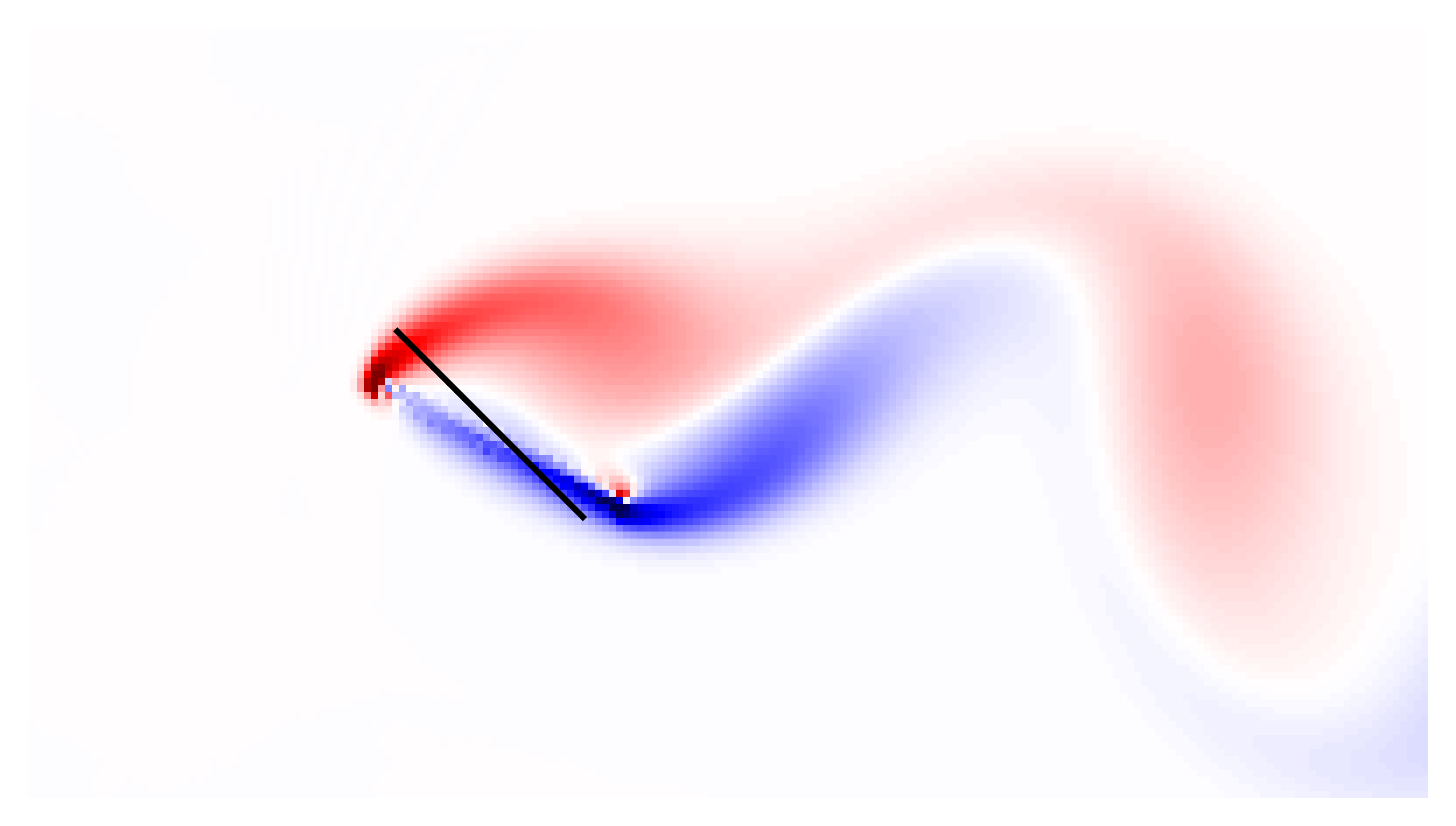}
        \caption{$t=3$}
        \label{fig:ppplate-4s}
    \end{subfigure}
    \caption{Pitch-plunge flat plate vortex dynamics for $\phi_h=90^{\circ}$ at 4 different instances of time over a single period. The black solid line represents a reference static plate without the pitching and plunging motions.}
    \label{fig:ppplate_snapshots}
\end{figure}

Figure~\ref{fig:ppplate_snapshots} presents representative flow snapshots of the pitch-plunge flat plate dynamics over a single period, illustrating the coupled motion and its influence on the resulting vortex structures. Since one period is approximately $3.9s$, the four snapshots shown here span the characteristic stages of the motion, highlighting how the plate traverses the fluid and how the associated vortex patterns are generated in its wake.

\begin{figure}
  \centering
  \includegraphics[width=0.7\linewidth]{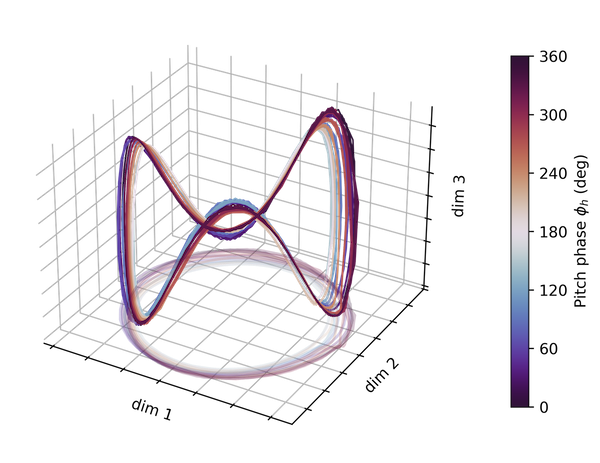}
  \caption{Isomap of 12 training trajectories of the pitch-plunge plate. }
  \label{fig:ppplate-isomap}
\end{figure}

For a better understanding of the intrinsic geometry, in Fig.~\ref{fig:ppplate-isomap}, the 12 high-dimensional training trajectories are embedded into a 3-dimensional representation using the Isomap algorithm \cite{tenenbaum2000global} to provide an intuition of the intrinsic geometry. Each trajectory, associated with one phase angle $\phi_h$, forms a loop that is topologically equivalent to $S^1$. All trajectories appear to lie on a skewed annulus $2D$ submanifold. However, due to the sparse sampling in phase angle, there are notable gaps among trajectories, which may pose a challenge to the data-driven models.

\paragraph{ResDMD.} The ResDMD is one type of the Dynamic Mode Decomposition method \cite{colbrook2024rigorous}.  The original ResDMD algorithm requires constructing the full order transition matrix of size $n\times n$, where $n$ is the ambient dimension.  However, this is infeasible in this high-dimensional setting as $n=179101$.  Therefore, based on the truncation order used for KRR $r=81$, we use $2r$ order PCA truncation to construct the transition matrix using the training trajectory data, and then the most accurate eigenpairs are retained for prediction.  In this section, the model retaining $2r$ modes is referred to as the ``full-order'' model.  The accuracy of the eigenpairs is quantified using a residual criterion, computed based on the validation data.  The hyperparameter in ResDMD is the number of eigenpairs to retain, which we refer to as the truncation order $\kappa$.

A validation study using the validation data is performed to select the truncation order over the range $\kappa \in\{9,11,13,…,162\}$.  
For this type of system, there is always one mean flow mode and a complex conjugate pair, so $\kappa$ should be an odd number and the increment of $\kappa$ is $2$.
For each $\kappa$, we quantify the prediction error using WNRMSE as in the KRR models. 
Eventually $\kappa=145$ corresponds to the lowest WNRMSE, and thus chosen as the hyperparameter for the ResDMD model for testing.

\begin{figure}[t]
    \centering
    \begin{subfigure}[b]{0.4\textwidth}
        \centering
        \includegraphics[width=\linewidth]{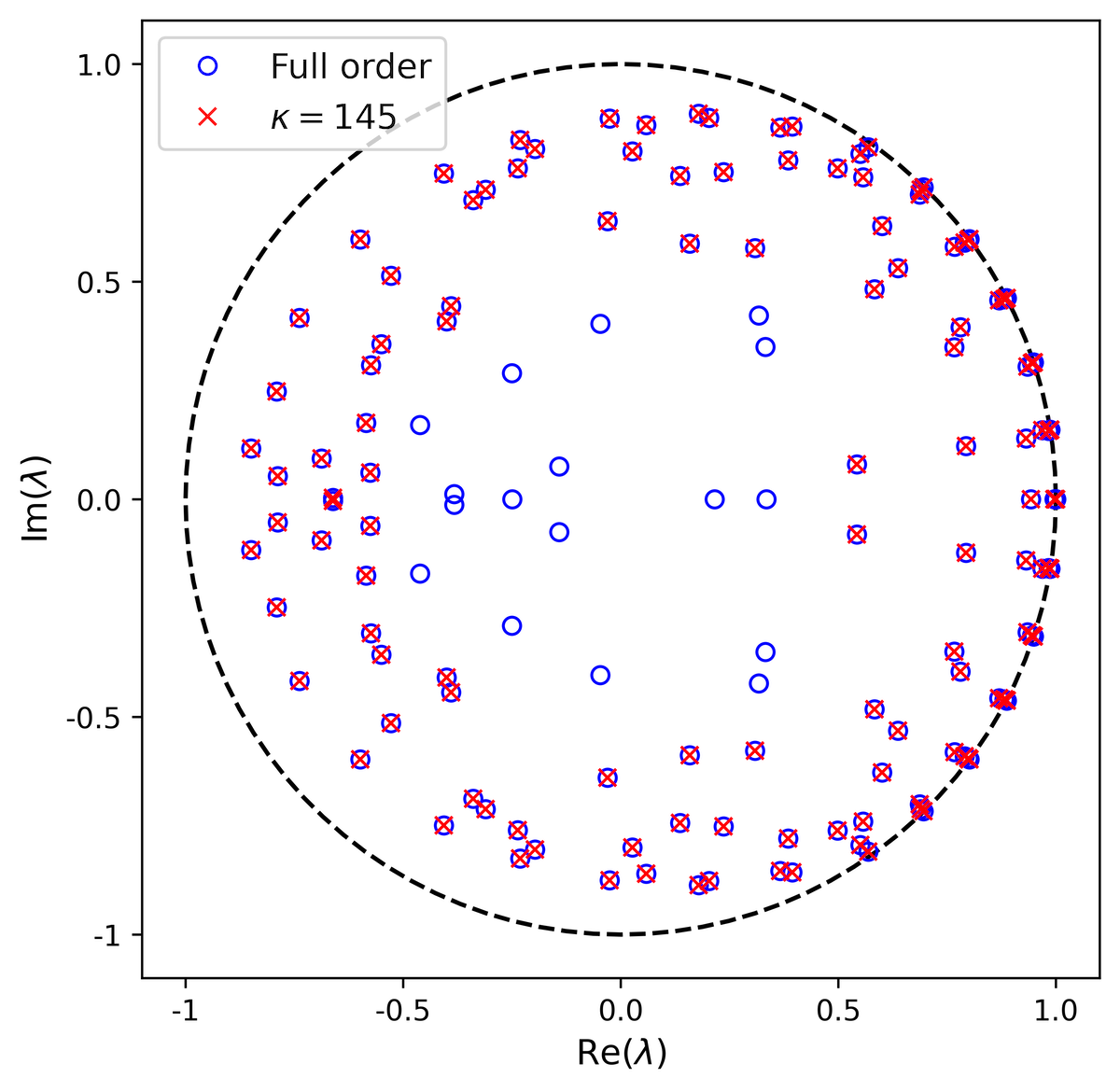}
        \caption{Full complex plane}
        \label{fig:ppplate-spectrum-full}
    \end{subfigure}
    \begin{subfigure}[b]{0.4\textwidth}
        \centering
        \includegraphics[width=\linewidth]{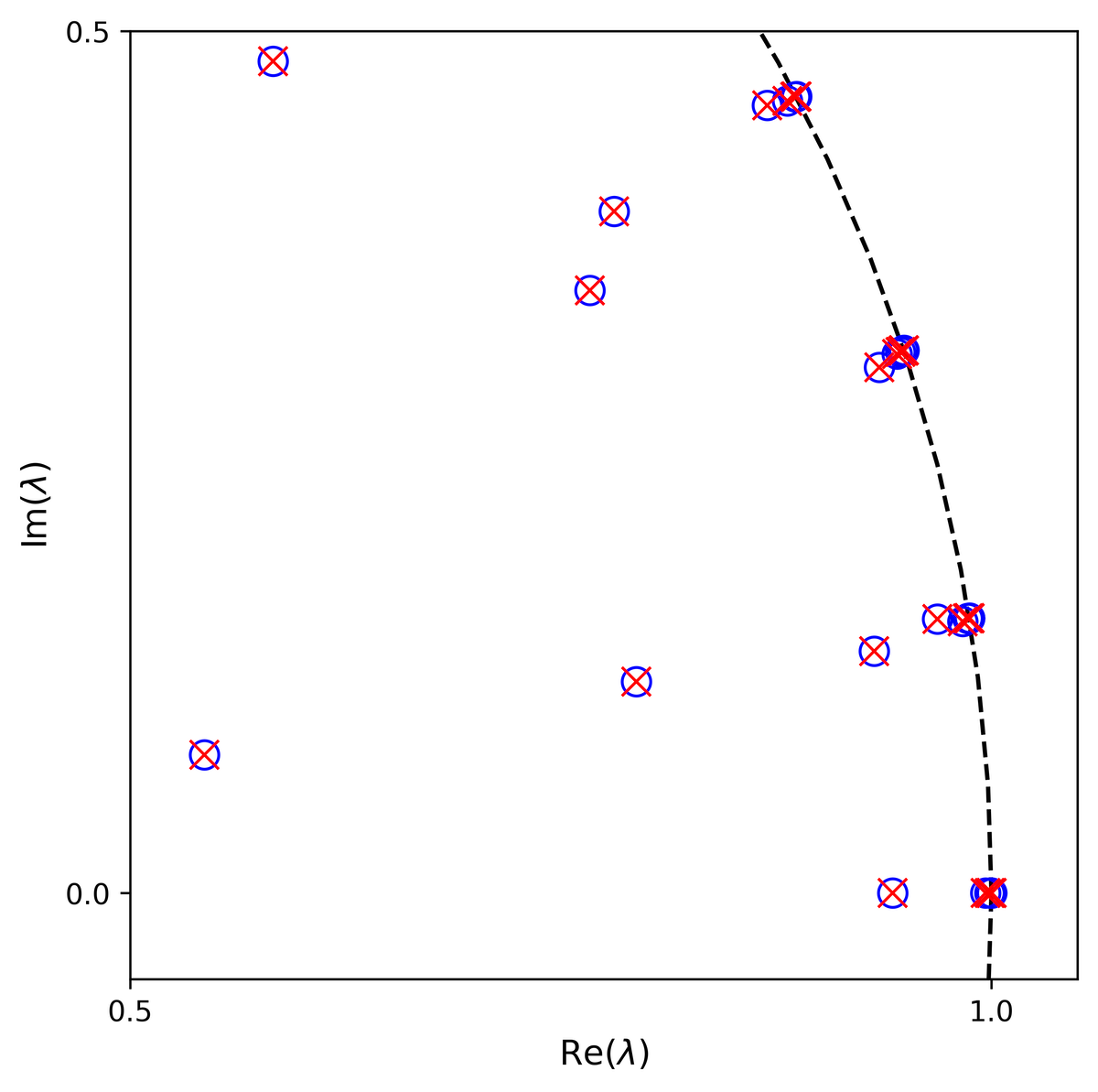}
        \caption{Zoomed in to the upper right side.}
        \label{fig:ppplate-spectrum-zoomin}
    \end{subfigure}
    \caption{The discrete-time eigenvalues of full order ($\kappa=162$) and optimal order ($\kappa=145$). The dashed line represents a unit circle in the complex plane.}
    \label{fig:ppplate-spectrum}
\end{figure}

In Fig.~\ref{fig:ppplate-spectrum-full}, the eigenvalues of the DMD models are shown. The blue hollow circles represent the eigenvalues of the full-order model whereas the red cross marks represent the eigenvalues of the optimal reduced order model.  For this type of periodic system, the eigenvalues are supposed to lie on the unit circle, representing simple harmonic modes.  However, because of the numerical errors, the eigenvalues are polluted and thus include many spurious ones that lie inside the circle, representing decaying modes.
The residual filtering eliminates the most spurious eigenvalues that are located near the center of the circle, but it is still not able to remove the outer eigenvalues.

We also examine the eigenvalues that appear to lie on the unit circle in Fig.~\ref{fig:ppplate-spectrum-zoomin}.  These eigenvalues are critical in prediction as they are associated with the dominant system frequency and corresponding modes \cite{song2025modal}.  Clearly, the eigenvalues are also slightly off from the unit circle, indicating slowly decaying modes, which contribute to the increasing error in the long-term prediction of ResDMD.

\paragraph{More test results.}
We provide DM and ResDMD predictions at three different instances of time, $t=\{3.0,42.2,77.5\}$ to illustrate the spatial distribution of error and its growth in time.  The RBF is not shown here as it reaches significant error even in the beginning of the prediction horizon. The NRMSE of DM is initially greater than that of RBF at $t=3.0$.  After approximately $10$ periods of oscillations ($\Delta t\approx 39.2$), the NRMSE of DM is sustained, while the NRMSE of ResDMD becomes similar to that of DM.
After $9$ more periods of oscillations, the NRMSE of DM is still similar, but the NRMSE of ResDMD becomes almost $30\%$ greater than that of DM, and it is expected to keep increasing beyond the considered prediction horizon.

\begin{figure}[t]
    \centering
    \begin{subfigure}[b]{0.8\textwidth}
        \centering
        \includegraphics[width=\textwidth]{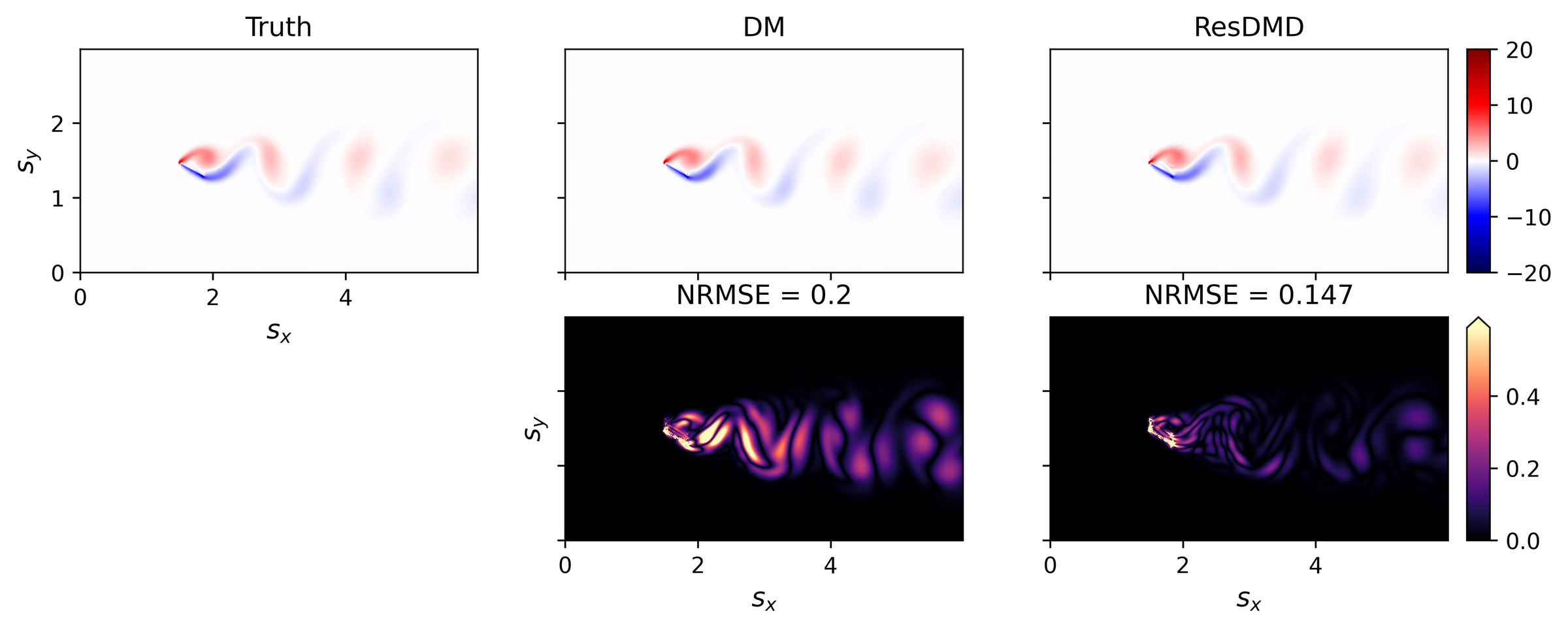}
        \caption{$t=3.0s$}
        \label{fig:ppplate-pred-3s}
    \end{subfigure}
    \hfill
    \begin{subfigure}[b]{0.8\textwidth}
        \centering
        \includegraphics[width=\textwidth]{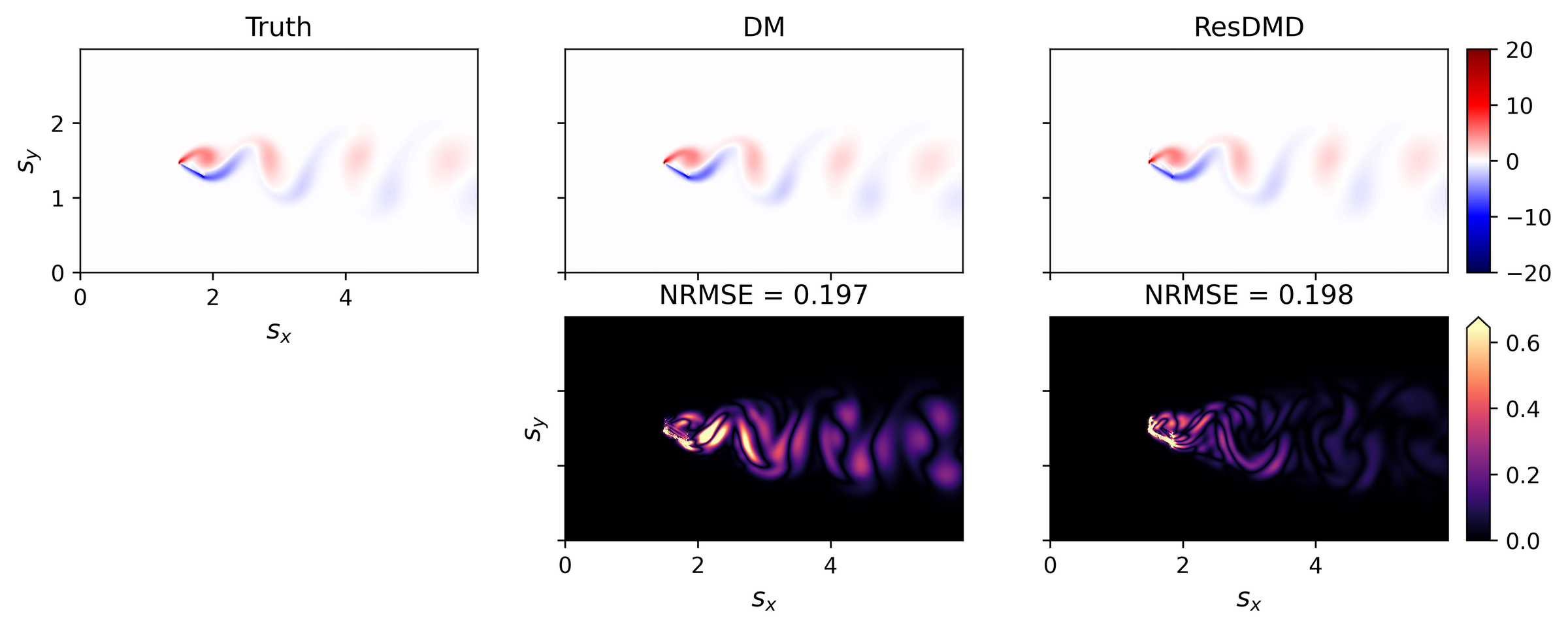}
        \caption{$t=42.2s$}
        \label{fig:ppplate-pred-42.2s}
    \end{subfigure}
    \hfill
    \begin{subfigure}[b]{0.8\textwidth}
        \centering
        \includegraphics[width=\textwidth]{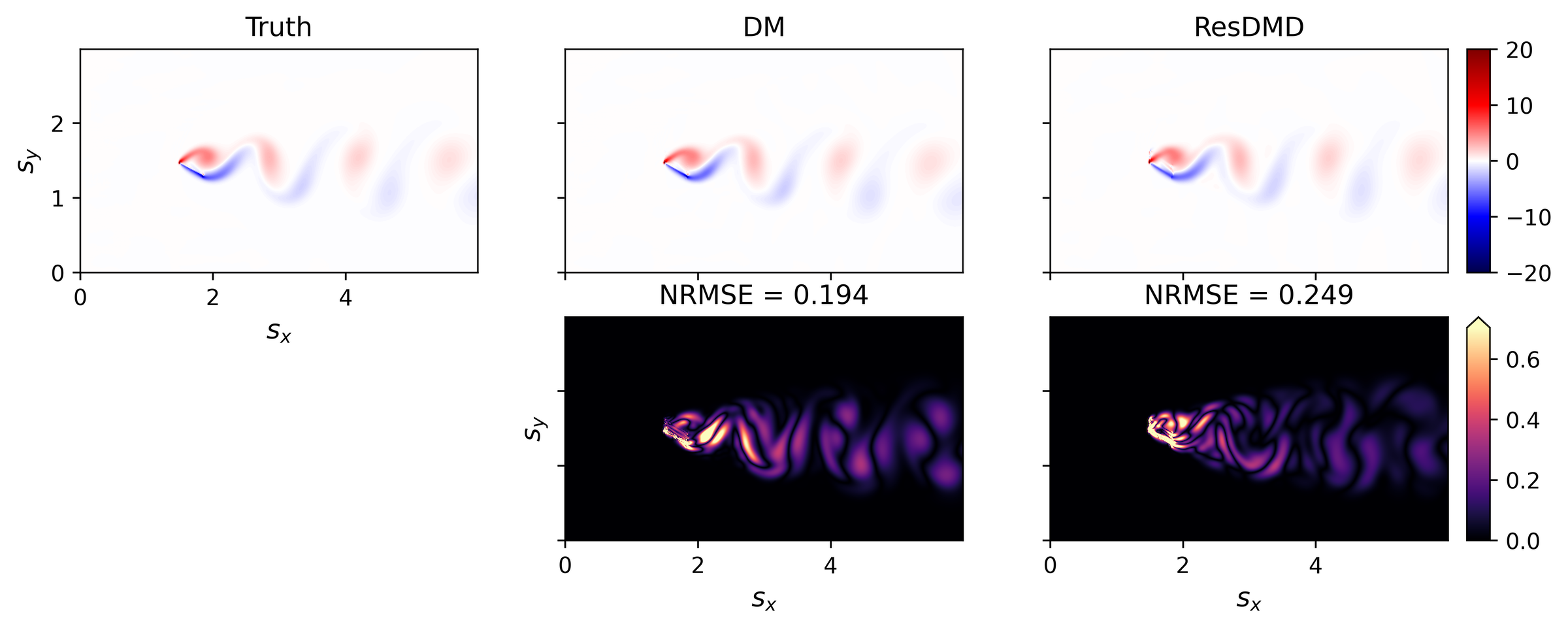}
        \caption{$t=77.5s$}
        \label{fig:ppplate-pred-77.5s}
    \end{subfigure}
    \caption{DM and ResDMD predictions and their errors at three different times $t=\{3.0,42.2,77.5\}s$.}
    \label{fig:ppplate_pred_snapshots}
\end{figure}

\end{document}